\pdfoutput=1

\documentclass[journal]{IEEEtran}
\usepackage{latexsym,graphicx,epsfig}
\usepackage{cite,subfig}
\usepackage{amsfonts}
\usepackage{amsmath}
\usepackage{amsthm}
\usepackage{amssymb}
\usepackage{amscd}
\usepackage[english]{babel}
\usepackage{simplemargins}
\usepackage{fancyhdr}
\usepackage{color}
\definecolor{AliceBlue}{rgb}{0.94,0.97,1.00}
\definecolor{AntiqueWhite1}{rgb}{1.00,0.94,0.86}
\definecolor{AntiqueWhite2}{rgb}{0.93,0.87,0.80}
\definecolor{AntiqueWhite3}{rgb}{0.80,0.75,0.69}
\definecolor{AntiqueWhite4}{rgb}{0.55,0.51,0.47}
\definecolor{AntiqueWhite}{rgb}{0.98,0.92,0.84}
\definecolor{BlanchedAlmond}{rgb}{1.00,0.92,0.80}
\definecolor{BlueViolet}{rgb}{0.54,0.17,0.89}
\definecolor{CadetBlue1}{rgb}{0.60,0.96,1.00}
\definecolor{CadetBlue2}{rgb}{0.56,0.90,0.93}
\definecolor{CadetBlue3}{rgb}{0.48,0.77,0.80}
\definecolor{CadetBlue4}{rgb}{0.33,0.53,0.55}
\definecolor{CadetBlue}{rgb}{0.37,0.62,0.63}
\definecolor{CornflowerBlue}{rgb}{0.39,0.58,0.93}
\definecolor{DarkBlue}{rgb}{0.00,0.00,0.55}
\definecolor{DarkCyan}{rgb}{0.00,0.55,0.55}
\definecolor{DarkGoldenrod1}{rgb}{1.00,0.73,0.06}
\definecolor{DarkGoldenrod2}{rgb}{0.93,0.68,0.05}
\definecolor{DarkGoldenrod3}{rgb}{0.80,0.58,0.05}
\definecolor{DarkGoldenrod4}{rgb}{0.55,0.40,0.03}
\definecolor{DarkGoldenrod}{rgb}{0.72,0.53,0.04}
\definecolor{DarkGray}{rgb}{0.66,0.66,0.66}
\definecolor{DarkGreen}{rgb}{0.00,0.39,0.00}
\definecolor{DarkGrey}{rgb}{0.66,0.66,0.66}
\definecolor{DarkKhaki}{rgb}{0.74,0.72,0.42}
\definecolor{DarkMagenta}{rgb}{0.55,0.00,0.55}
\definecolor{DarkOliveGreen1}{rgb}{0.79,1.00,0.44}
\definecolor{DarkOliveGreen2}{rgb}{0.74,0.93,0.41}
\definecolor{DarkOliveGreen3}{rgb}{0.64,0.80,0.35}
\definecolor{DarkOliveGreen4}{rgb}{0.43,0.55,0.24}
\definecolor{DarkOliveGreen}{rgb}{0.33,0.42,0.18}
\definecolor{DarkOrange1}{rgb}{1.00,0.50,0.00}
\definecolor{DarkOrange2}{rgb}{0.93,0.46,0.00}
\definecolor{DarkOrange3}{rgb}{0.80,0.40,0.00}
\definecolor{DarkOrange4}{rgb}{0.55,0.27,0.00}
\definecolor{DarkOrange}{rgb}{1.00,0.55,0.00}
\definecolor{DarkOrchid1}{rgb}{0.75,0.24,1.00}
\definecolor{DarkOrchid2}{rgb}{0.70,0.23,0.93}
\definecolor{DarkOrchid3}{rgb}{0.60,0.20,0.80}
\definecolor{DarkOrchid4}{rgb}{0.41,0.13,0.55}
\definecolor{DarkOrchid}{rgb}{0.60,0.20,0.80}
\definecolor{DarkRed}{rgb}{0.55,0.00,0.00}
\definecolor{DarkSalmon}{rgb}{0.91,0.59,0.48}
\definecolor{DarkSeaGreen1}{rgb}{0.76,1.00,0.76}
\definecolor{DarkSeaGreen2}{rgb}{0.71,0.93,0.71}
\definecolor{DarkSeaGreen3}{rgb}{0.61,0.80,0.61}
\definecolor{DarkSeaGreen4}{rgb}{0.41,0.55,0.41}
\definecolor{DarkSeaGreen}{rgb}{0.56,0.74,0.56}
\definecolor{DarkSlateBlue}{rgb}{0.28,0.24,0.55}
\definecolor{DarkSlateGray1}{rgb}{0.59,1.00,1.00}
\definecolor{DarkSlateGray2}{rgb}{0.55,0.93,0.93}
\definecolor{DarkSlateGray3}{rgb}{0.47,0.80,0.80}
\definecolor{DarkSlateGray4}{rgb}{0.32,0.55,0.55}
\definecolor{DarkSlateGray}{rgb}{0.18,0.31,0.31}
\definecolor{DarkSlateGrey}{rgb}{0.18,0.31,0.31}
\definecolor{DarkTurquoise}{rgb}{0.00,0.81,0.82}
\definecolor{DarkViolet}{rgb}{0.58,0.00,0.83}
\definecolor{DeepPink1}{rgb}{1.00,0.08,0.58}
\definecolor{DeepPink2}{rgb}{0.93,0.07,0.54}
\definecolor{DeepPink3}{rgb}{0.80,0.06,0.46}
\definecolor{DeepPink4}{rgb}{0.55,0.04,0.31}
\definecolor{DeepPink}{rgb}{1.00,0.08,0.58}
\definecolor{DeepSkyBlue1}{rgb}{0.00,0.75,1.00}
\definecolor{DeepSkyBlue2}{rgb}{0.00,0.70,0.93}
\definecolor{DeepSkyBlue3}{rgb}{0.00,0.60,0.80}
\definecolor{DeepSkyBlue4}{rgb}{0.00,0.41,0.55}
\definecolor{DeepSkyBlue}{rgb}{0.00,0.75,1.00}
\definecolor{DimGray}{rgb}{0.41,0.41,0.41}
\definecolor{DimGrey}{rgb}{0.41,0.41,0.41}
\definecolor{DodgerBlue1}{rgb}{0.12,0.56,1.00}
\definecolor{DodgerBlue2}{rgb}{0.11,0.53,0.93}
\definecolor{DodgerBlue3}{rgb}{0.09,0.45,0.80}
\definecolor{DodgerBlue4}{rgb}{0.06,0.31,0.55}
\definecolor{DodgerBlue}{rgb}{0.12,0.56,1.00}
\definecolor{FloralWhite}{rgb}{1.00,0.98,0.94}
\definecolor{ForestGreen}{rgb}{0.13,0.55,0.13}
\definecolor{GhostWhite}{rgb}{0.97,0.97,1.00}
\definecolor{GreenYellow}{rgb}{0.68,1.00,0.18}
\definecolor{HotPink1}{rgb}{1.00,0.43,0.71}
\definecolor{HotPink2}{rgb}{0.93,0.42,0.65}
\definecolor{HotPink3}{rgb}{0.80,0.38,0.56}
\definecolor{HotPink4}{rgb}{0.55,0.23,0.38}
\definecolor{HotPink}{rgb}{1.00,0.41,0.71}
\definecolor{IndianRed1}{rgb}{1.00,0.42,0.42}
\definecolor{IndianRed2}{rgb}{0.93,0.39,0.39}
\definecolor{IndianRed3}{rgb}{0.80,0.33,0.33}
\definecolor{IndianRed4}{rgb}{0.55,0.23,0.23}
\definecolor{IndianRed}{rgb}{0.80,0.36,0.36}
\definecolor{LavenderBlush1}{rgb}{1.00,0.94,0.96}
\definecolor{LavenderBlush2}{rgb}{0.93,0.88,0.90}
\definecolor{LavenderBlush3}{rgb}{0.80,0.76,0.77}
\definecolor{LavenderBlush4}{rgb}{0.55,0.51,0.53}
\definecolor{LavenderBlush}{rgb}{1.00,0.94,0.96}
\definecolor{LawnGreen}{rgb}{0.49,0.99,0.00}
\definecolor{LemonChiffon1}{rgb}{1.00,0.98,0.80}
\definecolor{LemonChiffon2}{rgb}{0.93,0.91,0.75}
\definecolor{LemonChiffon3}{rgb}{0.80,0.79,0.65}
\definecolor{LemonChiffon4}{rgb}{0.55,0.54,0.44}
\definecolor{LemonChiffon}{rgb}{1.00,0.98,0.80}
\definecolor{LightBlue1}{rgb}{0.75,0.94,1.00}
\definecolor{LightBlue2}{rgb}{0.70,0.87,0.93}
\definecolor{LightBlue3}{rgb}{0.60,0.75,0.80}
\definecolor{LightBlue4}{rgb}{0.41,0.51,0.55}
\definecolor{LightBlue}{rgb}{0.68,0.85,0.90}
\definecolor{LightCoral}{rgb}{0.94,0.50,0.50}
\definecolor{LightCyan1}{rgb}{0.88,1.00,1.00}
\definecolor{LightCyan2}{rgb}{0.82,0.93,0.93}
\definecolor{LightCyan3}{rgb}{0.71,0.80,0.80}
\definecolor{LightCyan4}{rgb}{0.48,0.55,0.55}
\definecolor{LightCyan}{rgb}{0.88,1.00,1.00}
\definecolor{LightGoldenrod1}{rgb}{1.00,0.93,0.55}
\definecolor{LightGoldenrod2}{rgb}{0.93,0.86,0.51}
\definecolor{LightGoldenrod3}{rgb}{0.80,0.75,0.44}
\definecolor{LightGoldenrod4}{rgb}{0.55,0.51,0.30}
\definecolor{LightGoldenrodYellow}{rgb}{0.98,0.98,0.82}
\definecolor{LightGoldenrod}{rgb}{0.93,0.87,0.51}
\definecolor{LightGray}{rgb}{0.83,0.83,0.83}
\definecolor{LightGreen}{rgb}{0.56,0.93,0.56}
\definecolor{LightGrey}{rgb}{0.83,0.83,0.83}
\definecolor{LightPink1}{rgb}{1.00,0.68,0.73}
\definecolor{LightPink2}{rgb}{0.93,0.64,0.68}
\definecolor{LightPink3}{rgb}{0.80,0.55,0.58}
\definecolor{LightPink4}{rgb}{0.55,0.37,0.40}
\definecolor{LightPink}{rgb}{1.00,0.71,0.76}
\definecolor{LightSalmon1}{rgb}{1.00,0.63,0.48}
\definecolor{LightSalmon2}{rgb}{0.93,0.58,0.45}
\definecolor{LightSalmon3}{rgb}{0.80,0.51,0.38}
\definecolor{LightSalmon4}{rgb}{0.55,0.34,0.26}
\definecolor{LightSalmon}{rgb}{1.00,0.63,0.48}
\definecolor{LightSeaGreen}{rgb}{0.13,0.70,0.67}
\definecolor{LightSkyBlue1}{rgb}{0.69,0.89,1.00}
\definecolor{LightSkyBlue2}{rgb}{0.64,0.83,0.93}
\definecolor{LightSkyBlue3}{rgb}{0.55,0.71,0.80}
\definecolor{LightSkyBlue4}{rgb}{0.38,0.48,0.55}
\definecolor{LightSkyBlue}{rgb}{0.53,0.81,0.98}
\definecolor{LightSlateBlue}{rgb}{0.52,0.44,1.00}
\definecolor{LightSlateGray}{rgb}{0.47,0.53,0.60}
\definecolor{LightSlateGrey}{rgb}{0.47,0.53,0.60}
\definecolor{LightSteelBlue1}{rgb}{0.79,0.88,1.00}
\definecolor{LightSteelBlue2}{rgb}{0.74,0.82,0.93}
\definecolor{LightSteelBlue3}{rgb}{0.64,0.71,0.80}
\definecolor{LightSteelBlue4}{rgb}{0.43,0.48,0.55}
\definecolor{LightSteelBlue}{rgb}{0.69,0.77,0.87}
\definecolor{LightYellow1}{rgb}{1.00,1.00,0.88}
\definecolor{LightYellow2}{rgb}{0.93,0.93,0.82}
\definecolor{LightYellow3}{rgb}{0.80,0.80,0.71}
\definecolor{LightYellow4}{rgb}{0.55,0.55,0.48}
\definecolor{LightYellow}{rgb}{1.00,1.00,0.88}
\definecolor{LimeGreen}{rgb}{0.20,0.80,0.20}
\definecolor{MediumAquamarine}{rgb}{0.40,0.80,0.67}
\definecolor{MediumBlue}{rgb}{0.00,0.00,0.80}
\definecolor{MediumOrchid1}{rgb}{0.88,0.40,1.00}
\definecolor{MediumOrchid2}{rgb}{0.82,0.37,0.93}
\definecolor{MediumOrchid3}{rgb}{0.71,0.32,0.80}
\definecolor{MediumOrchid4}{rgb}{0.48,0.22,0.55}
\definecolor{MediumOrchid}{rgb}{0.73,0.33,0.83}
\definecolor{MediumPurple1}{rgb}{0.67,0.51,1.00}
\definecolor{MediumPurple2}{rgb}{0.62,0.47,0.93}
\definecolor{MediumPurple3}{rgb}{0.54,0.41,0.80}
\definecolor{MediumPurple4}{rgb}{0.36,0.28,0.55}
\definecolor{MediumPurple}{rgb}{0.58,0.44,0.86}
\definecolor{MediumSeaGreen}{rgb}{0.24,0.70,0.44}
\definecolor{MediumSlateBlue}{rgb}{0.48,0.41,0.93}
\definecolor{MediumSpringGreen}{rgb}{0.00,0.98,0.60}
\definecolor{MediumTurquoise}{rgb}{0.28,0.82,0.80}
\definecolor{MediumVioletRed}{rgb}{0.78,0.08,0.52}
\definecolor{MidnightBlue}{rgb}{0.10,0.10,0.44}
\definecolor{MintCream}{rgb}{0.96,1.00,0.98}
\definecolor{MistyRose1}{rgb}{1.00,0.89,0.88}
\definecolor{MistyRose2}{rgb}{0.93,0.84,0.82}
\definecolor{MistyRose3}{rgb}{0.80,0.72,0.71}
\definecolor{MistyRose4}{rgb}{0.55,0.49,0.48}
\definecolor{MistyRose}{rgb}{1.00,0.89,0.88}
\definecolor{NavajoWhite1}{rgb}{1.00,0.87,0.68}
\definecolor{NavajoWhite2}{rgb}{0.93,0.81,0.63}
\definecolor{NavajoWhite3}{rgb}{0.80,0.70,0.55}
\definecolor{NavajoWhite4}{rgb}{0.55,0.47,0.37}
\definecolor{NavajoWhite}{rgb}{1.00,0.87,0.68}
\definecolor{NavyBlue}{rgb}{0.00,0.00,0.50}
\definecolor{OldLace}{rgb}{0.99,0.96,0.90}
\definecolor{OliveDrab1}{rgb}{0.75,1.00,0.24}
\definecolor{OliveDrab2}{rgb}{0.70,0.93,0.23}
\definecolor{OliveDrab3}{rgb}{0.60,0.80,0.20}
\definecolor{OliveDrab4}{rgb}{0.41,0.55,0.13}
\definecolor{OliveDrab}{rgb}{0.42,0.56,0.14}
\definecolor{OrangeRed1}{rgb}{1.00,0.27,0.00}
\definecolor{OrangeRed2}{rgb}{0.93,0.25,0.00}
\definecolor{OrangeRed3}{rgb}{0.80,0.22,0.00}
\definecolor{OrangeRed4}{rgb}{0.55,0.15,0.00}
\definecolor{OrangeRed}{rgb}{1.00,0.27,0.00}
\definecolor{PaleGoldenrod}{rgb}{0.93,0.91,0.67}
\definecolor{PaleGreen1}{rgb}{0.60,1.00,0.60}
\definecolor{PaleGreen2}{rgb}{0.56,0.93,0.56}
\definecolor{PaleGreen3}{rgb}{0.49,0.80,0.49}
\definecolor{PaleGreen4}{rgb}{0.33,0.55,0.33}
\definecolor{PaleGreen}{rgb}{0.60,0.98,0.60}
\definecolor{PaleTurquoise1}{rgb}{0.73,1.00,1.00}
\definecolor{PaleTurquoise2}{rgb}{0.68,0.93,0.93}
\definecolor{PaleTurquoise3}{rgb}{0.59,0.80,0.80}
\definecolor{PaleTurquoise4}{rgb}{0.40,0.55,0.55}
\definecolor{PaleTurquoise}{rgb}{0.69,0.93,0.93}
\definecolor{PaleVioletRed1}{rgb}{1.00,0.51,0.67}
\definecolor{PaleVioletRed2}{rgb}{0.93,0.47,0.62}
\definecolor{PaleVioletRed3}{rgb}{0.80,0.41,0.54}
\definecolor{PaleVioletRed4}{rgb}{0.55,0.28,0.36}
\definecolor{PaleVioletRed}{rgb}{0.86,0.44,0.58}
\definecolor{PapayaWhip}{rgb}{1.00,0.94,0.84}
\definecolor{PeachPuff1}{rgb}{1.00,0.85,0.73}
\definecolor{PeachPuff2}{rgb}{0.93,0.80,0.68}
\definecolor{PeachPuff3}{rgb}{0.80,0.69,0.58}
\definecolor{PeachPuff4}{rgb}{0.55,0.47,0.40}
\definecolor{PeachPuff}{rgb}{1.00,0.85,0.73}
\definecolor{PowderBlue}{rgb}{0.69,0.88,0.90}
\definecolor{RosyBrown1}{rgb}{1.00,0.76,0.76}
\definecolor{RosyBrown2}{rgb}{0.93,0.71,0.71}
\definecolor{RosyBrown3}{rgb}{0.80,0.61,0.61}
\definecolor{RosyBrown4}{rgb}{0.55,0.41,0.41}
\definecolor{RosyBrown}{rgb}{0.74,0.56,0.56}
\definecolor{RoyalBlue1}{rgb}{0.28,0.46,1.00}
\definecolor{RoyalBlue2}{rgb}{0.26,0.43,0.93}
\definecolor{RoyalBlue3}{rgb}{0.23,0.37,0.80}
\definecolor{RoyalBlue4}{rgb}{0.15,0.25,0.55}
\definecolor{RoyalBlue}{rgb}{0.25,0.41,0.88}
\definecolor{SaddleBrown}{rgb}{0.55,0.27,0.07}
\definecolor{SandyBrown}{rgb}{0.96,0.64,0.38}
\definecolor{SeaGreen1}{rgb}{0.33,1.00,0.62}
\definecolor{SeaGreen2}{rgb}{0.31,0.93,0.58}
\definecolor{SeaGreen3}{rgb}{0.26,0.80,0.50}
\definecolor{SeaGreen4}{rgb}{0.18,0.55,0.34}
\definecolor{SeaGreen}{rgb}{0.18,0.55,0.34}
\definecolor{SkyBlue1}{rgb}{0.53,0.81,1.00}
\definecolor{SkyBlue2}{rgb}{0.49,0.75,0.93}
\definecolor{SkyBlue3}{rgb}{0.42,0.65,0.80}
\definecolor{SkyBlue4}{rgb}{0.29,0.44,0.55}
\definecolor{SkyBlue}{rgb}{0.53,0.81,0.92}
\definecolor{SlateBlue1}{rgb}{0.51,0.44,1.00}
\definecolor{SlateBlue2}{rgb}{0.48,0.40,0.93}
\definecolor{SlateBlue3}{rgb}{0.41,0.35,0.80}
\definecolor{SlateBlue4}{rgb}{0.28,0.24,0.55}
\definecolor{SlateBlue}{rgb}{0.42,0.35,0.80}
\definecolor{SlateGray1}{rgb}{0.78,0.89,1.00}
\definecolor{SlateGray2}{rgb}{0.73,0.83,0.93}
\definecolor{SlateGray3}{rgb}{0.62,0.71,0.80}
\definecolor{SlateGray4}{rgb}{0.42,0.48,0.55}
\definecolor{SlateGray}{rgb}{0.44,0.50,0.56}
\definecolor{SlateGrey}{rgb}{0.44,0.50,0.56}
\definecolor{SpringGreen1}{rgb}{0.00,1.00,0.50}
\definecolor{SpringGreen2}{rgb}{0.00,0.93,0.46}
\definecolor{SpringGreen3}{rgb}{0.00,0.80,0.40}
\definecolor{SpringGreen4}{rgb}{0.00,0.55,0.27}
\definecolor{SpringGreen}{rgb}{0.00,1.00,0.50}
\definecolor{SteelBlue1}{rgb}{0.39,0.72,1.00}
\definecolor{SteelBlue2}{rgb}{0.36,0.67,0.93}
\definecolor{SteelBlue3}{rgb}{0.31,0.58,0.80}
\definecolor{SteelBlue4}{rgb}{0.21,0.39,0.55}
\definecolor{SteelBlue}{rgb}{0.27,0.51,0.71}
\definecolor{VioletRed1}{rgb}{1.00,0.24,0.59}
\definecolor{VioletRed2}{rgb}{0.93,0.23,0.55}
\definecolor{VioletRed3}{rgb}{0.80,0.20,0.47}
\definecolor{VioletRed4}{rgb}{0.55,0.13,0.32}
\definecolor{VioletRed}{rgb}{0.82,0.13,0.56}
\definecolor{WhiteSmoke}{rgb}{0.96,0.96,0.96}
\definecolor{YellowGreen}{rgb}{0.60,0.80,0.20}
\definecolor{aliceblue}{rgb}{0.94,0.97,1.00}
\definecolor{antiquewhite}{rgb}{0.98,0.92,0.84}
\definecolor{aquamarine1}{rgb}{0.50,1.00,0.83}
\definecolor{aquamarine2}{rgb}{0.46,0.93,0.78}
\definecolor{aquamarine3}{rgb}{0.40,0.80,0.67}
\definecolor{aquamarine4}{rgb}{0.27,0.55,0.45}
\definecolor{aquamarine}{rgb}{0.50,1.00,0.83}
\definecolor{azure1}{rgb}{0.94,1.00,1.00}
\definecolor{azure2}{rgb}{0.88,0.93,0.93}
\definecolor{azure3}{rgb}{0.76,0.80,0.80}
\definecolor{azure4}{rgb}{0.51,0.55,0.55}
\definecolor{azure}{rgb}{0.94,1.00,1.00}
\definecolor{beige}{rgb}{0.96,0.96,0.86}
\definecolor{bisque1}{rgb}{1.00,0.89,0.77}
\definecolor{bisque2}{rgb}{0.93,0.84,0.72}
\definecolor{bisque3}{rgb}{0.80,0.72,0.62}
\definecolor{bisque4}{rgb}{0.55,0.49,0.42}
\definecolor{bisque}{rgb}{1.00,0.89,0.77}
\definecolor{black}{rgb}{0.00,0.00,0.00}
\definecolor{blanchedalmond}{rgb}{1.00,0.92,0.80}
\definecolor{blue1}{rgb}{0.00,0.00,1.00}
\definecolor{blue2}{rgb}{0.00,0.00,0.93}
\definecolor{blue3}{rgb}{0.00,0.00,0.80}
\definecolor{blue4}{rgb}{0.00,0.00,0.55}
\definecolor{blueviolet}{rgb}{0.54,0.17,0.89}
\definecolor{blue}{rgb}{0.00,0.00,1.00}
\definecolor{brown1}{rgb}{1.00,0.25,0.25}
\definecolor{brown2}{rgb}{0.93,0.23,0.23}
\definecolor{brown3}{rgb}{0.80,0.20,0.20}
\definecolor{brown4}{rgb}{0.55,0.14,0.14}
\definecolor{brown}{rgb}{0.65,0.16,0.16}
\definecolor{burlywood1}{rgb}{1.00,0.83,0.61}
\definecolor{burlywood2}{rgb}{0.93,0.77,0.57}
\definecolor{burlywood3}{rgb}{0.80,0.67,0.49}
\definecolor{burlywood4}{rgb}{0.55,0.45,0.33}
\definecolor{burlywood}{rgb}{0.87,0.72,0.53}
\definecolor{cadetblue}{rgb}{0.37,0.62,0.63}
\definecolor{chartreuse1}{rgb}{0.50,1.00,0.00}
\definecolor{chartreuse2}{rgb}{0.46,0.93,0.00}
\definecolor{chartreuse3}{rgb}{0.40,0.80,0.00}
\definecolor{chartreuse4}{rgb}{0.27,0.55,0.00}
\definecolor{chartreuse}{rgb}{0.50,1.00,0.00}
\definecolor{chocolate1}{rgb}{1.00,0.50,0.14}
\definecolor{chocolate2}{rgb}{0.93,0.46,0.13}
\definecolor{chocolate3}{rgb}{0.80,0.40,0.11}
\definecolor{chocolate4}{rgb}{0.55,0.27,0.07}
\definecolor{chocolate}{rgb}{0.82,0.41,0.12}
\definecolor{coral1}{rgb}{1.00,0.45,0.34}
\definecolor{coral2}{rgb}{0.93,0.42,0.31}
\definecolor{coral3}{rgb}{0.80,0.36,0.27}
\definecolor{coral4}{rgb}{0.55,0.24,0.18}
\definecolor{coral}{rgb}{1.00,0.50,0.31}
\definecolor{cornflowerblue}{rgb}{0.39,0.58,0.93}
\definecolor{cornsilk1}{rgb}{1.00,0.97,0.86}
\definecolor{cornsilk2}{rgb}{0.93,0.91,0.80}
\definecolor{cornsilk3}{rgb}{0.80,0.78,0.69}
\definecolor{cornsilk4}{rgb}{0.55,0.53,0.47}
\definecolor{cornsilk}{rgb}{1.00,0.97,0.86}
\definecolor{cyan1}{rgb}{0.00,1.00,1.00}
\definecolor{cyan2}{rgb}{0.00,0.93,0.93}
\definecolor{cyan3}{rgb}{0.00,0.80,0.80}
\definecolor{cyan4}{rgb}{0.00,0.55,0.55}
\definecolor{cyan}{rgb}{0.00,1.00,1.00}
\definecolor{darkblue}{rgb}{0.00,0.00,0.55}
\definecolor{darkcyan}{rgb}{0.00,0.55,0.55}
\definecolor{darkgoldenrod}{rgb}{0.72,0.53,0.04}
\definecolor{darkgray}{rgb}{0.66,0.66,0.66}
\definecolor{darkgreen}{rgb}{0.00,0.39,0.00}
\definecolor{darkgrey}{rgb}{0.66,0.66,0.66}
\definecolor{darkkhaki}{rgb}{0.74,0.72,0.42}
\definecolor{darkmagenta}{rgb}{0.55,0.00,0.55}
\definecolor{darkolive}{rgb}{0.33,0.42,0.18}
\definecolor{darkorange}{rgb}{1.00,0.55,0.00}
\definecolor{darkorchid}{rgb}{0.60,0.20,0.80}
\definecolor{darkred}{rgb}{0.55,0.00,0.00}
\definecolor{darksalmon}{rgb}{0.91,0.59,0.48}
\definecolor{darksea}{rgb}{0.56,0.74,0.56}
\definecolor{darkslate}{rgb}{0.18,0.31,0.31}
\definecolor{darkslate}{rgb}{0.18,0.31,0.31}
\definecolor{darkslate}{rgb}{0.28,0.24,0.55}
\definecolor{darkturquoise}{rgb}{0.00,0.81,0.82}
\definecolor{darkviolet}{rgb}{0.58,0.00,0.83}
\definecolor{deeppink}{rgb}{1.00,0.08,0.58}
\definecolor{deepsky}{rgb}{0.00,0.75,1.00}
\definecolor{dimgray}{rgb}{0.41,0.41,0.41}
\definecolor{dimgrey}{rgb}{0.41,0.41,0.41}
\definecolor{dodgerblue}{rgb}{0.12,0.56,1.00}
\definecolor{firebrick1}{rgb}{1.00,0.19,0.19}
\definecolor{firebrick2}{rgb}{0.93,0.17,0.17}
\definecolor{firebrick3}{rgb}{0.80,0.15,0.15}
\definecolor{firebrick4}{rgb}{0.55,0.10,0.10}
\definecolor{firebrick}{rgb}{0.70,0.13,0.13}
\definecolor{floralwhite}{rgb}{1.00,0.98,0.94}
\definecolor{forestgreen}{rgb}{0.13,0.55,0.13}
\definecolor{gainsboro}{rgb}{0.86,0.86,0.86}
\definecolor{ghostwhite}{rgb}{0.97,0.97,1.00}
\definecolor{gold1}{rgb}{1.00,0.84,0.00}
\definecolor{gold2}{rgb}{0.93,0.79,0.00}
\definecolor{gold3}{rgb}{0.80,0.68,0.00}
\definecolor{gold4}{rgb}{0.55,0.46,0.00}
\definecolor{goldenrod1}{rgb}{1.00,0.76,0.15}
\definecolor{goldenrod2}{rgb}{0.93,0.71,0.13}
\definecolor{goldenrod3}{rgb}{0.80,0.61,0.11}
\definecolor{goldenrod4}{rgb}{0.55,0.41,0.08}
\definecolor{goldenrod}{rgb}{0.85,0.65,0.13}
\definecolor{gold}{rgb}{1.00,0.84,0.00}
\definecolor{gray0}{rgb}{0.00,0.00,0.00}
\definecolor{gray100}{rgb}{1.00,1.00,1.00}
\definecolor{gray10}{rgb}{0.10,0.10,0.10}
\definecolor{gray11}{rgb}{0.11,0.11,0.11}
\definecolor{gray12}{rgb}{0.12,0.12,0.12}
\definecolor{gray13}{rgb}{0.13,0.13,0.13}
\definecolor{gray14}{rgb}{0.14,0.14,0.14}
\definecolor{gray15}{rgb}{0.15,0.15,0.15}
\definecolor{gray16}{rgb}{0.16,0.16,0.16}
\definecolor{gray17}{rgb}{0.17,0.17,0.17}
\definecolor{gray18}{rgb}{0.18,0.18,0.18}
\definecolor{gray19}{rgb}{0.19,0.19,0.19}
\definecolor{gray1}{rgb}{0.01,0.01,0.01}
\definecolor{gray20}{rgb}{0.20,0.20,0.20}
\definecolor{gray21}{rgb}{0.21,0.21,0.21}
\definecolor{gray22}{rgb}{0.22,0.22,0.22}
\definecolor{gray23}{rgb}{0.23,0.23,0.23}
\definecolor{gray24}{rgb}{0.24,0.24,0.24}
\definecolor{gray25}{rgb}{0.25,0.25,0.25}
\definecolor{gray26}{rgb}{0.26,0.26,0.26}
\definecolor{gray27}{rgb}{0.27,0.27,0.27}
\definecolor{gray28}{rgb}{0.28,0.28,0.28}
\definecolor{gray29}{rgb}{0.29,0.29,0.29}
\definecolor{gray2}{rgb}{0.02,0.02,0.02}
\definecolor{gray30}{rgb}{0.30,0.30,0.30}
\definecolor{gray31}{rgb}{0.31,0.31,0.31}
\definecolor{gray32}{rgb}{0.32,0.32,0.32}
\definecolor{gray33}{rgb}{0.33,0.33,0.33}
\definecolor{gray34}{rgb}{0.34,0.34,0.34}
\definecolor{gray35}{rgb}{0.35,0.35,0.35}
\definecolor{gray36}{rgb}{0.36,0.36,0.36}
\definecolor{gray37}{rgb}{0.37,0.37,0.37}
\definecolor{gray38}{rgb}{0.38,0.38,0.38}
\definecolor{gray39}{rgb}{0.39,0.39,0.39}
\definecolor{gray3}{rgb}{0.03,0.03,0.03}
\definecolor{gray40}{rgb}{0.40,0.40,0.40}
\definecolor{gray41}{rgb}{0.41,0.41,0.41}
\definecolor{gray42}{rgb}{0.42,0.42,0.42}
\definecolor{gray43}{rgb}{0.43,0.43,0.43}
\definecolor{gray44}{rgb}{0.44,0.44,0.44}
\definecolor{gray45}{rgb}{0.45,0.45,0.45}
\definecolor{gray46}{rgb}{0.46,0.46,0.46}
\definecolor{gray47}{rgb}{0.47,0.47,0.47}
\definecolor{gray48}{rgb}{0.48,0.48,0.48}
\definecolor{gray49}{rgb}{0.49,0.49,0.49}
\definecolor{gray4}{rgb}{0.04,0.04,0.04}
\definecolor{gray50}{rgb}{0.50,0.50,0.50}
\definecolor{gray51}{rgb}{0.51,0.51,0.51}
\definecolor{gray52}{rgb}{0.52,0.52,0.52}
\definecolor{gray53}{rgb}{0.53,0.53,0.53}
\definecolor{gray54}{rgb}{0.54,0.54,0.54}
\definecolor{gray55}{rgb}{0.55,0.55,0.55}
\definecolor{gray56}{rgb}{0.56,0.56,0.56}
\definecolor{gray57}{rgb}{0.57,0.57,0.57}
\definecolor{gray58}{rgb}{0.58,0.58,0.58}
\definecolor{gray59}{rgb}{0.59,0.59,0.59}
\definecolor{gray5}{rgb}{0.05,0.05,0.05}
\definecolor{gray60}{rgb}{0.60,0.60,0.60}
\definecolor{gray61}{rgb}{0.61,0.61,0.61}
\definecolor{gray62}{rgb}{0.62,0.62,0.62}
\definecolor{gray63}{rgb}{0.63,0.63,0.63}
\definecolor{gray64}{rgb}{0.64,0.64,0.64}
\definecolor{gray65}{rgb}{0.65,0.65,0.65}
\definecolor{gray66}{rgb}{0.66,0.66,0.66}
\definecolor{gray67}{rgb}{0.67,0.67,0.67}
\definecolor{gray68}{rgb}{0.68,0.68,0.68}
\definecolor{gray69}{rgb}{0.69,0.69,0.69}
\definecolor{gray6}{rgb}{0.06,0.06,0.06}
\definecolor{gray70}{rgb}{0.70,0.70,0.70}
\definecolor{gray71}{rgb}{0.71,0.71,0.71}
\definecolor{gray72}{rgb}{0.72,0.72,0.72}
\definecolor{gray73}{rgb}{0.73,0.73,0.73}
\definecolor{gray74}{rgb}{0.74,0.74,0.74}
\definecolor{gray75}{rgb}{0.75,0.75,0.75}
\definecolor{gray76}{rgb}{0.76,0.76,0.76}
\definecolor{gray77}{rgb}{0.77,0.77,0.77}
\definecolor{gray78}{rgb}{0.78,0.78,0.78}
\definecolor{gray79}{rgb}{0.79,0.79,0.79}
\definecolor{gray7}{rgb}{0.07,0.07,0.07}
\definecolor{gray80}{rgb}{0.80,0.80,0.80}
\definecolor{gray81}{rgb}{0.81,0.81,0.81}
\definecolor{gray82}{rgb}{0.82,0.82,0.82}
\definecolor{gray83}{rgb}{0.83,0.83,0.83}
\definecolor{gray84}{rgb}{0.84,0.84,0.84}
\definecolor{gray85}{rgb}{0.85,0.85,0.85}
\definecolor{gray86}{rgb}{0.86,0.86,0.86}
\definecolor{gray87}{rgb}{0.87,0.87,0.87}
\definecolor{gray88}{rgb}{0.88,0.88,0.88}
\definecolor{gray89}{rgb}{0.89,0.89,0.89}
\definecolor{gray8}{rgb}{0.08,0.08,0.08}
\definecolor{gray90}{rgb}{0.90,0.90,0.90}
\definecolor{gray91}{rgb}{0.91,0.91,0.91}
\definecolor{gray92}{rgb}{0.92,0.92,0.92}
\definecolor{gray93}{rgb}{0.93,0.93,0.93}
\definecolor{gray94}{rgb}{0.94,0.94,0.94}
\definecolor{gray95}{rgb}{0.95,0.95,0.95}
\definecolor{gray96}{rgb}{0.96,0.96,0.96}
\definecolor{gray97}{rgb}{0.97,0.97,0.97}
\definecolor{gray98}{rgb}{0.98,0.98,0.98}
\definecolor{gray99}{rgb}{0.99,0.99,0.99}
\definecolor{gray9}{rgb}{0.09,0.09,0.09}
\definecolor{gray}{rgb}{0.75,0.75,0.75}
\definecolor{green1}{rgb}{0.00,1.00,0.00}
\definecolor{green2}{rgb}{0.00,0.93,0.00}
\definecolor{green3}{rgb}{0.00,0.80,0.00}
\definecolor{green4}{rgb}{0.00,0.55,0.00}
\definecolor{greenyellow}{rgb}{0.68,1.00,0.18}
\definecolor{green}{rgb}{0.00,1.00,0.00}
\definecolor{grey0}{rgb}{0.00,0.00,0.00}
\definecolor{grey100}{rgb}{1.00,1.00,1.00}
\definecolor{grey10}{rgb}{0.10,0.10,0.10}
\definecolor{grey11}{rgb}{0.11,0.11,0.11}
\definecolor{grey12}{rgb}{0.12,0.12,0.12}
\definecolor{grey13}{rgb}{0.13,0.13,0.13}
\definecolor{grey14}{rgb}{0.14,0.14,0.14}
\definecolor{grey15}{rgb}{0.15,0.15,0.15}
\definecolor{grey16}{rgb}{0.16,0.16,0.16}
\definecolor{grey17}{rgb}{0.17,0.17,0.17}
\definecolor{grey18}{rgb}{0.18,0.18,0.18}
\definecolor{grey19}{rgb}{0.19,0.19,0.19}
\definecolor{grey1}{rgb}{0.01,0.01,0.01}
\definecolor{grey20}{rgb}{0.20,0.20,0.20}
\definecolor{grey21}{rgb}{0.21,0.21,0.21}
\definecolor{grey22}{rgb}{0.22,0.22,0.22}
\definecolor{grey23}{rgb}{0.23,0.23,0.23}
\definecolor{grey24}{rgb}{0.24,0.24,0.24}
\definecolor{grey25}{rgb}{0.25,0.25,0.25}
\definecolor{grey26}{rgb}{0.26,0.26,0.26}
\definecolor{grey27}{rgb}{0.27,0.27,0.27}
\definecolor{grey28}{rgb}{0.28,0.28,0.28}
\definecolor{grey29}{rgb}{0.29,0.29,0.29}
\definecolor{grey2}{rgb}{0.02,0.02,0.02}
\definecolor{grey30}{rgb}{0.30,0.30,0.30}
\definecolor{grey31}{rgb}{0.31,0.31,0.31}
\definecolor{grey32}{rgb}{0.32,0.32,0.32}
\definecolor{grey33}{rgb}{0.33,0.33,0.33}
\definecolor{grey34}{rgb}{0.34,0.34,0.34}
\definecolor{grey35}{rgb}{0.35,0.35,0.35}
\definecolor{grey36}{rgb}{0.36,0.36,0.36}
\definecolor{grey37}{rgb}{0.37,0.37,0.37}
\definecolor{grey38}{rgb}{0.38,0.38,0.38}
\definecolor{grey39}{rgb}{0.39,0.39,0.39}
\definecolor{grey3}{rgb}{0.03,0.03,0.03}
\definecolor{grey40}{rgb}{0.40,0.40,0.40}
\definecolor{grey41}{rgb}{0.41,0.41,0.41}
\definecolor{grey42}{rgb}{0.42,0.42,0.42}
\definecolor{grey43}{rgb}{0.43,0.43,0.43}
\definecolor{grey44}{rgb}{0.44,0.44,0.44}
\definecolor{grey45}{rgb}{0.45,0.45,0.45}
\definecolor{grey46}{rgb}{0.46,0.46,0.46}
\definecolor{grey47}{rgb}{0.47,0.47,0.47}
\definecolor{grey48}{rgb}{0.48,0.48,0.48}
\definecolor{grey49}{rgb}{0.49,0.49,0.49}
\definecolor{grey4}{rgb}{0.04,0.04,0.04}
\definecolor{grey50}{rgb}{0.50,0.50,0.50}
\definecolor{grey51}{rgb}{0.51,0.51,0.51}
\definecolor{grey52}{rgb}{0.52,0.52,0.52}
\definecolor{grey53}{rgb}{0.53,0.53,0.53}
\definecolor{grey54}{rgb}{0.54,0.54,0.54}
\definecolor{grey55}{rgb}{0.55,0.55,0.55}
\definecolor{grey56}{rgb}{0.56,0.56,0.56}
\definecolor{grey57}{rgb}{0.57,0.57,0.57}
\definecolor{grey58}{rgb}{0.58,0.58,0.58}
\definecolor{grey59}{rgb}{0.59,0.59,0.59}
\definecolor{grey5}{rgb}{0.05,0.05,0.05}
\definecolor{grey60}{rgb}{0.60,0.60,0.60}
\definecolor{grey61}{rgb}{0.61,0.61,0.61}
\definecolor{grey62}{rgb}{0.62,0.62,0.62}
\definecolor{grey63}{rgb}{0.63,0.63,0.63}
\definecolor{grey64}{rgb}{0.64,0.64,0.64}
\definecolor{grey65}{rgb}{0.65,0.65,0.65}
\definecolor{grey66}{rgb}{0.66,0.66,0.66}
\definecolor{grey67}{rgb}{0.67,0.67,0.67}
\definecolor{grey68}{rgb}{0.68,0.68,0.68}
\definecolor{grey69}{rgb}{0.69,0.69,0.69}
\definecolor{grey6}{rgb}{0.06,0.06,0.06}
\definecolor{grey70}{rgb}{0.70,0.70,0.70}
\definecolor{grey71}{rgb}{0.71,0.71,0.71}
\definecolor{grey72}{rgb}{0.72,0.72,0.72}
\definecolor{grey73}{rgb}{0.73,0.73,0.73}
\definecolor{grey74}{rgb}{0.74,0.74,0.74}
\definecolor{grey75}{rgb}{0.75,0.75,0.75}
\definecolor{grey76}{rgb}{0.76,0.76,0.76}
\definecolor{grey77}{rgb}{0.77,0.77,0.77}
\definecolor{grey78}{rgb}{0.78,0.78,0.78}
\definecolor{grey79}{rgb}{0.79,0.79,0.79}
\definecolor{grey7}{rgb}{0.07,0.07,0.07}
\definecolor{grey80}{rgb}{0.80,0.80,0.80}
\definecolor{grey81}{rgb}{0.81,0.81,0.81}
\definecolor{grey82}{rgb}{0.82,0.82,0.82}
\definecolor{grey83}{rgb}{0.83,0.83,0.83}
\definecolor{grey84}{rgb}{0.84,0.84,0.84}
\definecolor{grey85}{rgb}{0.85,0.85,0.85}
\definecolor{grey86}{rgb}{0.86,0.86,0.86}
\definecolor{grey87}{rgb}{0.87,0.87,0.87}
\definecolor{grey88}{rgb}{0.88,0.88,0.88}
\definecolor{grey89}{rgb}{0.89,0.89,0.89}
\definecolor{grey8}{rgb}{0.08,0.08,0.08}
\definecolor{grey90}{rgb}{0.90,0.90,0.90}
\definecolor{grey91}{rgb}{0.91,0.91,0.91}
\definecolor{grey92}{rgb}{0.92,0.92,0.92}
\definecolor{grey93}{rgb}{0.93,0.93,0.93}
\definecolor{grey94}{rgb}{0.94,0.94,0.94}
\definecolor{grey95}{rgb}{0.95,0.95,0.95}
\definecolor{grey96}{rgb}{0.96,0.96,0.96}
\definecolor{grey97}{rgb}{0.97,0.97,0.97}
\definecolor{grey98}{rgb}{0.98,0.98,0.98}
\definecolor{grey99}{rgb}{0.99,0.99,0.99}
\definecolor{grey9}{rgb}{0.09,0.09,0.09}
\definecolor{grey}{rgb}{0.75,0.75,0.75}
\definecolor{honeydew1}{rgb}{0.94,1.00,0.94}
\definecolor{honeydew2}{rgb}{0.88,0.93,0.88}
\definecolor{honeydew3}{rgb}{0.76,0.80,0.76}
\definecolor{honeydew4}{rgb}{0.51,0.55,0.51}
\definecolor{honeydew}{rgb}{0.94,1.00,0.94}
\definecolor{hotpink}{rgb}{1.00,0.41,0.71}
\definecolor{indianred}{rgb}{0.80,0.36,0.36}
\definecolor{ivory1}{rgb}{1.00,1.00,0.94}
\definecolor{ivory2}{rgb}{0.93,0.93,0.88}
\definecolor{ivory3}{rgb}{0.80,0.80,0.76}
\definecolor{ivory4}{rgb}{0.55,0.55,0.51}
\definecolor{ivory}{rgb}{1.00,1.00,0.94}
\definecolor{khaki1}{rgb}{1.00,0.96,0.56}
\definecolor{khaki2}{rgb}{0.93,0.90,0.52}
\definecolor{khaki3}{rgb}{0.80,0.78,0.45}
\definecolor{khaki4}{rgb}{0.55,0.53,0.31}
\definecolor{khaki}{rgb}{0.94,0.90,0.55}
\definecolor{lavenderblush}{rgb}{1.00,0.94,0.96}
\definecolor{lavender}{rgb}{0.90,0.90,0.98}
\definecolor{lawngreen}{rgb}{0.49,0.99,0.00}
\definecolor{lemonchiffon}{rgb}{1.00,0.98,0.80}
\definecolor{lightblue}{rgb}{0.68,0.85,0.90}
\definecolor{lightcoral}{rgb}{0.94,0.50,0.50}
\definecolor{lightcyan}{rgb}{0.88,1.00,1.00}
\definecolor{lightgoldenrod}{rgb}{0.93,0.87,0.51}
\definecolor{lightgoldenrod}{rgb}{0.98,0.98,0.82}
\definecolor{lightgray}{rgb}{0.83,0.83,0.83}
\definecolor{lightgreen}{rgb}{0.56,0.93,0.56}
\definecolor{lightgrey}{rgb}{0.83,0.83,0.83}
\definecolor{lightpink}{rgb}{1.00,0.71,0.76}
\definecolor{lightsalmon}{rgb}{1.00,0.63,0.48}
\definecolor{lightsea}{rgb}{0.13,0.70,0.67}
\definecolor{lightsky}{rgb}{0.53,0.81,0.98}
\definecolor{lightslate}{rgb}{0.47,0.53,0.60}
\definecolor{lightslate}{rgb}{0.47,0.53,0.60}
\definecolor{lightslate}{rgb}{0.52,0.44,1.00}
\definecolor{lightsteel}{rgb}{0.69,0.77,0.87}
\definecolor{lightyellow}{rgb}{1.00,1.00,0.88}
\definecolor{limegreen}{rgb}{0.20,0.80,0.20}
\definecolor{linen}{rgb}{0.98,0.94,0.90}
\definecolor{magenta1}{rgb}{1.00,0.00,1.00}
\definecolor{magenta2}{rgb}{0.93,0.00,0.93}
\definecolor{magenta3}{rgb}{0.80,0.00,0.80}
\definecolor{magenta4}{rgb}{0.55,0.00,0.55}
\definecolor{magenta}{rgb}{1.00,0.00,1.00}
\definecolor{maroon1}{rgb}{1.00,0.20,0.70}
\definecolor{maroon2}{rgb}{0.93,0.19,0.65}
\definecolor{maroon3}{rgb}{0.80,0.16,0.56}
\definecolor{maroon4}{rgb}{0.55,0.11,0.38}
\definecolor{maroon}{rgb}{0.69,0.19,0.38}
\definecolor{mediumaquamarine}{rgb}{0.40,0.80,0.67}
\definecolor{mediumblue}{rgb}{0.00,0.00,0.80}
\definecolor{mediumorchid}{rgb}{0.73,0.33,0.83}
\definecolor{mediumpurple}{rgb}{0.58,0.44,0.86}
\definecolor{mediumsea}{rgb}{0.24,0.70,0.44}
\definecolor{mediumslate}{rgb}{0.48,0.41,0.93}
\definecolor{mediumspring}{rgb}{0.00,0.98,0.60}
\definecolor{mediumturquoise}{rgb}{0.28,0.82,0.80}
\definecolor{mediumviolet}{rgb}{0.78,0.08,0.52}
\definecolor{midnightblue}{rgb}{0.10,0.10,0.44}
\definecolor{mintcream}{rgb}{0.96,1.00,0.98}
\definecolor{mistyrose}{rgb}{1.00,0.89,0.88}
\definecolor{moccasin}{rgb}{1.00,0.89,0.71}
\definecolor{navajowhite}{rgb}{1.00,0.87,0.68}
\definecolor{navyblue}{rgb}{0.00,0.00,0.50}
\definecolor{navy}{rgb}{0.00,0.00,0.50}
\definecolor{oldlace}{rgb}{0.99,0.96,0.90}
\definecolor{olivedrab}{rgb}{0.42,0.56,0.14}
\definecolor{orange1}{rgb}{1.00,0.65,0.00}
\definecolor{orange2}{rgb}{0.93,0.60,0.00}
\definecolor{orange3}{rgb}{0.80,0.52,0.00}
\definecolor{orange4}{rgb}{0.55,0.35,0.00}
\definecolor{orangered}{rgb}{1.00,0.27,0.00}
\definecolor{orange}{rgb}{1.00,0.65,0.00}
\definecolor{orchid1}{rgb}{1.00,0.51,0.98}
\definecolor{orchid2}{rgb}{0.93,0.48,0.91}
\definecolor{orchid3}{rgb}{0.80,0.41,0.79}
\definecolor{orchid4}{rgb}{0.55,0.28,0.54}
\definecolor{orchid}{rgb}{0.85,0.44,0.84}
\definecolor{palegoldenrod}{rgb}{0.93,0.91,0.67}
\definecolor{palegreen}{rgb}{0.60,0.98,0.60}
\definecolor{paleturquoise}{rgb}{0.69,0.93,0.93}
\definecolor{paleviolet}{rgb}{0.86,0.44,0.58}
\definecolor{papayawhip}{rgb}{1.00,0.94,0.84}
\definecolor{peachpuff}{rgb}{1.00,0.85,0.73}
\definecolor{peru}{rgb}{0.80,0.52,0.25}
\definecolor{pink1}{rgb}{1.00,0.71,0.77}
\definecolor{pink2}{rgb}{0.93,0.66,0.72}
\definecolor{pink3}{rgb}{0.80,0.57,0.62}
\definecolor{pink4}{rgb}{0.55,0.39,0.42}
\definecolor{pink}{rgb}{1.00,0.75,0.80}
\definecolor{plum1}{rgb}{1.00,0.73,1.00}
\definecolor{plum2}{rgb}{0.93,0.68,0.93}
\definecolor{plum3}{rgb}{0.80,0.59,0.80}
\definecolor{plum4}{rgb}{0.55,0.40,0.55}
\definecolor{plum}{rgb}{0.87,0.63,0.87}
\definecolor{powderblue}{rgb}{0.69,0.88,0.90}
\definecolor{purple1}{rgb}{0.61,0.19,1.00}
\definecolor{purple2}{rgb}{0.57,0.17,0.93}
\definecolor{purple3}{rgb}{0.49,0.15,0.80}
\definecolor{purple4}{rgb}{0.33,0.10,0.55}
\definecolor{purple}{rgb}{0.63,0.13,0.94}
\definecolor{red1}{rgb}{1.00,0.00,0.00}
\definecolor{red2}{rgb}{0.93,0.00,0.00}
\definecolor{red3}{rgb}{0.80,0.00,0.00}
\definecolor{red4}{rgb}{0.55,0.00,0.00}
\definecolor{red}{rgb}{1.00,0.00,0.00}
\definecolor{rosybrown}{rgb}{0.74,0.56,0.56}
\definecolor{royalblue}{rgb}{0.25,0.41,0.88}
\definecolor{saddlebrown}{rgb}{0.55,0.27,0.07}
\definecolor{salmon1}{rgb}{1.00,0.55,0.41}
\definecolor{salmon2}{rgb}{0.93,0.51,0.38}
\definecolor{salmon3}{rgb}{0.80,0.44,0.33}
\definecolor{salmon4}{rgb}{0.55,0.30,0.22}
\definecolor{salmon}{rgb}{0.98,0.50,0.45}
\definecolor{sandybrown}{rgb}{0.96,0.64,0.38}
\definecolor{seagreen}{rgb}{0.18,0.55,0.34}
\definecolor{seashell1}{rgb}{1.00,0.96,0.93}
\definecolor{seashell2}{rgb}{0.93,0.90,0.87}
\definecolor{seashell3}{rgb}{0.80,0.77,0.75}
\definecolor{seashell4}{rgb}{0.55,0.53,0.51}
\definecolor{seashell}{rgb}{1.00,0.96,0.93}
\definecolor{sienna1}{rgb}{1.00,0.51,0.28}
\definecolor{sienna2}{rgb}{0.93,0.47,0.26}
\definecolor{sienna3}{rgb}{0.80,0.41,0.22}
\definecolor{sienna4}{rgb}{0.55,0.28,0.15}
\definecolor{sienna}{rgb}{0.63,0.32,0.18}
\definecolor{skyblue}{rgb}{0.53,0.81,0.92}
\definecolor{slateblue}{rgb}{0.42,0.35,0.80}
\definecolor{slategray}{rgb}{0.44,0.50,0.56}
\definecolor{slategrey}{rgb}{0.44,0.50,0.56}
\definecolor{snow1}{rgb}{1.00,0.98,0.98}
\definecolor{snow2}{rgb}{0.93,0.91,0.91}
\definecolor{snow3}{rgb}{0.80,0.79,0.79}
\definecolor{snow4}{rgb}{0.55,0.54,0.54}
\definecolor{snow}{rgb}{1.00,0.98,0.98}
\definecolor{springgreen}{rgb}{0.00,1.00,0.50}
\definecolor{steelblue}{rgb}{0.27,0.51,0.71}
\definecolor{tan1}{rgb}{1.00,0.65,0.31}
\definecolor{tan2}{rgb}{0.93,0.60,0.29}
\definecolor{tan3}{rgb}{0.80,0.52,0.25}
\definecolor{tan4}{rgb}{0.55,0.35,0.17}
\definecolor{tan}{rgb}{0.82,0.71,0.55}
\definecolor{thistle1}{rgb}{1.00,0.88,1.00}
\definecolor{thistle2}{rgb}{0.93,0.82,0.93}
\definecolor{thistle3}{rgb}{0.80,0.71,0.80}
\definecolor{thistle4}{rgb}{0.55,0.48,0.55}
\definecolor{thistle}{rgb}{0.85,0.75,0.85}
\definecolor{tomato1}{rgb}{1.00,0.39,0.28}
\definecolor{tomato2}{rgb}{0.93,0.36,0.26}
\definecolor{tomato3}{rgb}{0.80,0.31,0.22}
\definecolor{tomato4}{rgb}{0.55,0.21,0.15}
\definecolor{tomato}{rgb}{1.00,0.39,0.28}
\definecolor{turquoise1}{rgb}{0.00,0.96,1.00}
\definecolor{turquoise2}{rgb}{0.00,0.90,0.93}
\definecolor{turquoise3}{rgb}{0.00,0.77,0.80}
\definecolor{turquoise4}{rgb}{0.00,0.53,0.55}
\definecolor{turquoise}{rgb}{0.25,0.88,0.82}
\definecolor{violetred}{rgb}{0.82,0.13,0.56}
\definecolor{violet}{rgb}{0.93,0.51,0.93}
\definecolor{wheat1}{rgb}{1.00,0.91,0.73}
\definecolor{wheat2}{rgb}{0.93,0.85,0.68}
\definecolor{wheat3}{rgb}{0.80,0.73,0.59}
\definecolor{wheat4}{rgb}{0.55,0.49,0.40}
\definecolor{wheat}{rgb}{0.96,0.87,0.70}
\definecolor{whitesmoke}{rgb}{0.96,0.96,0.96}
\definecolor{white}{rgb}{1.00,1.00,1.00}
\definecolor{yellow1}{rgb}{1.00,1.00,0.00}
\definecolor{yellow2}{rgb}{0.93,0.93,0.00}
\definecolor{yellow3}{rgb}{0.80,0.80,0.00}
\definecolor{yellow4}{rgb}{0.55,0.55,0.00}
\definecolor{yellowgreen}{rgb}{0.60,0.80,0.20}
\definecolor{yellow}{rgb}{1.00,1.00,0.00}

\newcommand{\mt}{\mathbf}   % German
\newcommand{\mc}{\mathcal}  % special
\newcommand{\mb}{\mathbb}   % same as special

\numberwithin{equation}{section}
\begin{document}
\title{Spatially-Adaptive Reconstruction in Computed Tomography using Neural Networks}
\author{Joseph Shtok, Michael Zibulevsky
         and Michael Elad, Fellow, IEEE.
        \thanks{All authors are with the Computer Science Department,
        Technion - Israel Institute of Technology, Israel.}
        \thanks
        {The research leading to these results has received funding from the European Research Council under European Union's Seventh Framework Program, ERC Grant agreement no. 320649. This research was supported by Gurwin Family Fund.}
}
\maketitle

\begin{abstract}
We propose a supervised machine learning approach for boosting existing signal and image recovery  methods and demonstrate its efficacy on example of image reconstruction in computed tomography. Our technique is based on a local nonlinear fusion of several image estimates, all obtained by applying a chosen reconstruction algorithm with different values of its control parameters. Usually such output images have  different bias/variance trade-off. The fusion of the images is performed by  feed-forward neural network trained on a set of known examples. Numerical experiments show an improvement in reconstruction quality relatively to existing direct and iterative reconstruction methods.
\end{abstract}

\begin{IEEEkeywords}
Computed Tomography, Low-Dose Reconstruction, Neural Networks, Supervised Learning, Filtered-Back-Projection (FBP).
\end{IEEEkeywords}

%==========================================================================================================================
%==========================================================================================================================

\section{Introduction}
Computed tomography (CT) imaging produces an attenuation map of the scanned object, by sequentially irradiating it with X-rays from several directions. The integral attenuation of the X-rays, measured by comparing the radiation intensity entering and leaving the body, forms the raw data for the CT imaging. In practice, these photon count measurements are degraded by stochastic noise, typically modeled as instances of Poisson random variables. There are also other degradation effects due to a number of physical phenomena -- see {\em e.g.} \cite{RiBi06} for a detailed account.

Given the projection data, known as the sinogram, a reconstruction process can be performed in order to recover the attenuation map. Various such algorithms exist, ranging from the simple and still very popular Filtered-Back-Projection (FBP) \cite{ShKr78}, and all the way to the more advanced Bayesian-inspired iterative algorithms (see e.g., \cite{ElFe02,WaLi06b}) that take the statistical nature of the measurements and the unknown image into account. Since CT relies on X-ray, which is an ionizing radiation known to be dangerous to living tissues, there is a dire and constant need to improve the reconstruction algorithms in an attempt to enable reduction of radiation dose.

In this work we are concerned with the question of image post-processing, following the CT reconstruction, for the purpose of getting better quality CT image, thereby permitting an eventual radiation-dose reduction. The proposed method  does not focus on a specific CT reconstruction algorithm, nor the properties of the images it produces. Instead, we take a generic approach which adapts, in an off-line learning process, to any such given algorithm. The only requirement is the access to design parameters of the reconstruction procedure which influence the nature of the output image, such as the resolution-variance trade-off.

We aim to exploit the fact that any reconstruction algorithm can provide more image information if instead of one fixed value of a parameter (or a vector of them) controlling the reconstruction, few different values are used (leading to different versions of the image). In order to extract this information from a collection of image versions, we use an Artificial Neural Network (ANN) \cite{Hayk94}. The proposed method can also use other techniques for computing a non-linear multivariate regression function.

Neural networks have been used extensively in medical imaging, particularly for the purpose of CT reconstruction (see Section \ref{sect:bkgnd-ANN} for an overview). Here we propose a new constellation, which consists in a local fusion of the different image versions, aimed at an improved reconstruction quality. We use a set of intensity values from a neighborhood of a pixel $q$, taken from the different versions, as inputs to the network, and train it to compute a (smaller) neighborhood of $q$ which values are as close as possible (in Mean-Squared-Error or other sense) to those found in the reference image. As we show in this paper, the proposed approach enables an improvement of the variance-resolution trade-off of a given reconstruction algorithm, i.e. producing images with a reduced amount of noise without compromising the spatial resolution and without introducing artifacts.

This paper is organized as follows: Sections \ref{sect:CT} and \ref{sect:bkgnd-ANN} are devoted to a brief and general discussion on CT scan/reconstruction and artificial neural networks. Readers familiar with these topics may skip and start reading at Section \ref{sect:RVTF-concept}, where the core concept of this work is detailed. This section also contains an illustration on one-dimensional piece-wise constant signals, where it is easy to appreciate the action of the proposed algorithm and the effect of local fusion performed by a neural network. In the sequel, the proposed method is implemented on two tomographic reconstruction methods: boosting the Filtered Back-Projection (FBP) is presented in Section \ref{sect:RVTF-FBP} and the same for Penalized Weighted Least-Squares (PWLS) method is described in Section \ref{sect:RVTF-PWLS-it}. We conclude this work by discussing the computational complexity of the proposed algorithms in Section \ref{sect:complex}, and a summary of this work and its potential implications in Section \ref{sect:summary}.

%==========================================================================================================================
%==========================================================================================================================

\section{Background on Computed Tomography} \label{sect:CT}

%==========================================================================================================================

\subsection{Mathematical Model of CT Scan} \label{sect:model-scan}

In the process of a CT scan, the object is radiated with X-rays. In this work we consider a reconstruction in a plane from rays incident only to this plane (the two-dimensional tomography). From the mathematical point of view, the considered object is a function $f(x)$ in the plane, which values are the attenuation coefficients of composing materials (i.e., tissues). When the measured photon counts are perfect, the measurements are directly related to the X-ray transform of the function $f(x)$ as a collection of all the straight lines passing through the object, and the value associated with each such line is the integral of $f(x)$ along it. In two dimensions, and under the assumption of a full rotated and parallel beam scan, this coincides with the Radon transform $\mt{R}f$.

Let $\ell$ be a straight line from an X-ray source to a detector. The ideal photon count $\lambda_{\ell}$, measured by the detector is related to $\mt{R}f$ via the function
\begin{equation}\label{eq:model1}
\lambda_{\ell} = \lambda_0e^{-[\mt{R}f]_{\ell}},
\end{equation}
where $\lambda_0$ is the blank scan count. The scanned data is stored in a matrix which columns correspond to the sampled angle $\theta$; each such column is referred to as a "view" or a "projection", and is acquired, schematically, by a parallel array of X-rays passing through the object at the corresponding angle. The rows of the matrix, corresponding to the sampled values of the distance $s$, are called the "bins" of each projection. According to the Equation (\ref{eq:model1}), for reconstruction purposes the measurements data undergoes the log transform
\begin{equation}\label{eq:g-y-connection}
g_{\ell}=-log(\dfrac{\lambda_{\ell}}{\lambda_0}).
\end{equation}
We refer to $g$ as the \textit{sinogram}. The name indicates that every point in the image space traces a sine curve in this domain. Since the sinogram matrix is the (sampled) Radon transform of the original image $f(x)$, a discrete version of the image can be reconstructed by applying the inverse Radon transform (see Section \ref{sect:BK-recon}).

Each measured photon count $y_{\ell}$ is typically interpreted as an instance of the random variable $Y_{\ell}$ following a Poisson distribution $ Y_{\ell} \sim Poisson(\lambda_{\ell}) $\cite{RiBi06,ThBo06,Bors09}. This reflects the photon count statistics at the detectors \cite{Hans81}. For a random variable $X\sim Poisson(\lambda)$, the standard deviation $\sigma_X$ satisfies $\sigma_X=\sqrt{\mb{E}(X)}$, and therefore the signal-to-noise ratio of $X$, $SNR(X)=\mb{E}(X)/\sigma_X = \sqrt{\mb{E}(X)}$ monotonously increases with its expected value.

\begin{figure}[htbp]
\centering
\includegraphics[width=1\columnwidth]{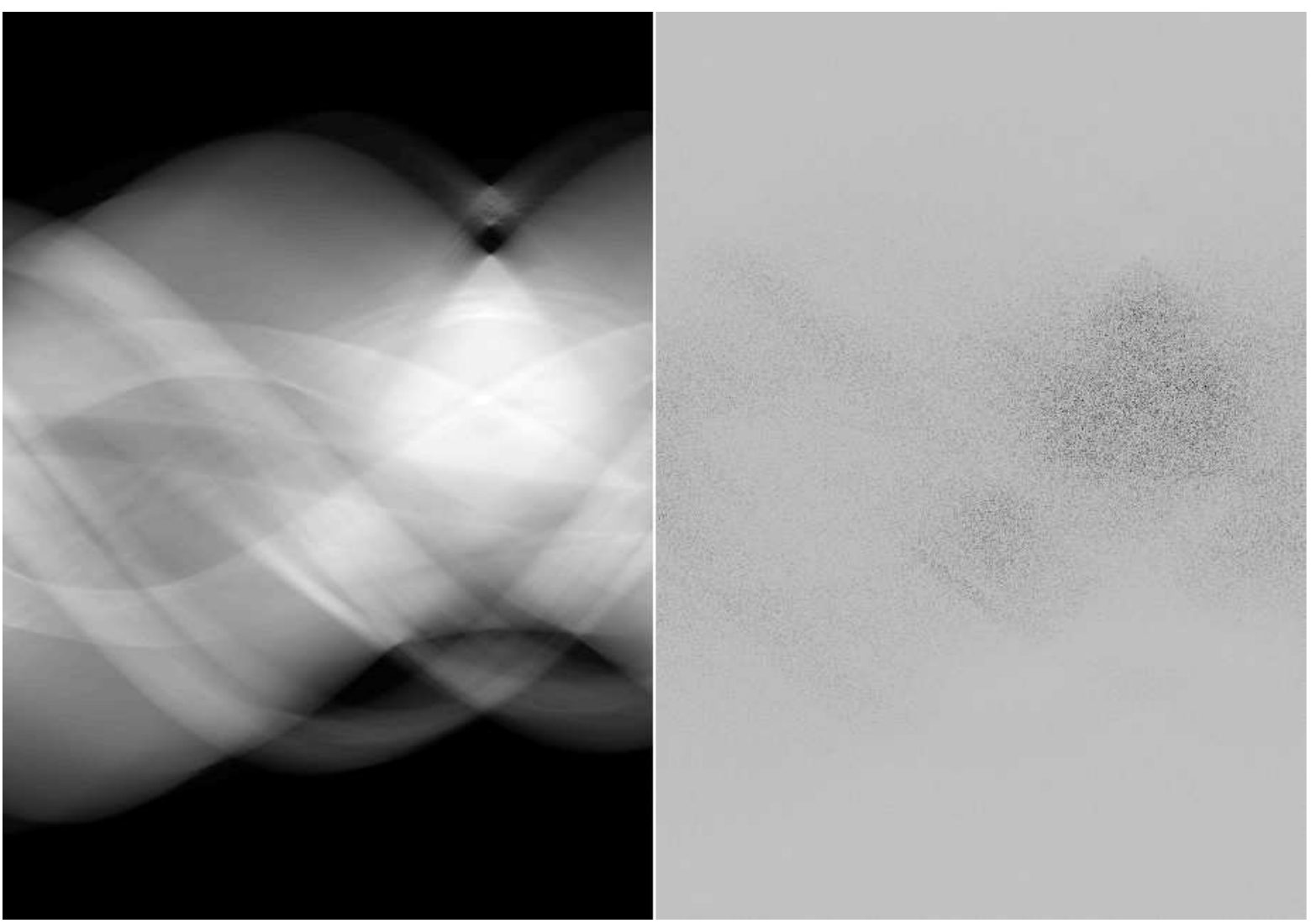} \\
\includegraphics[width=1\columnwidth]{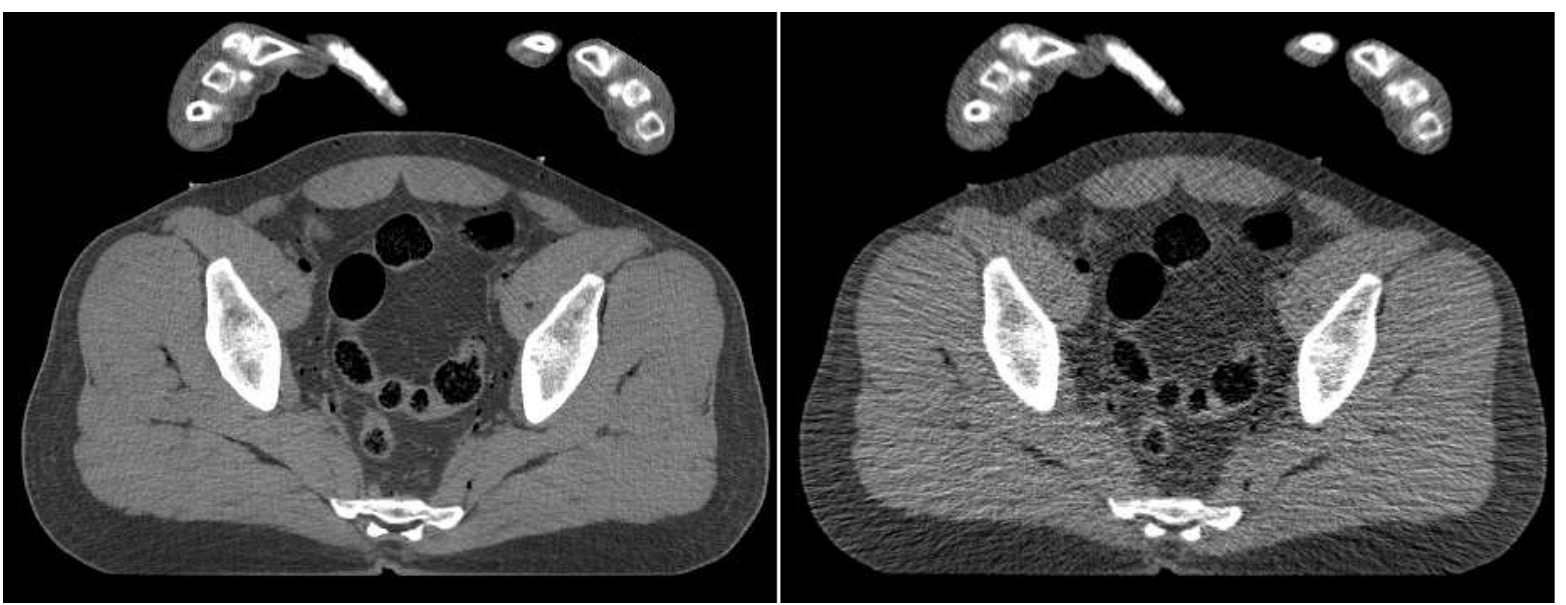}
\caption{Upper row, left to right: Exact sinogram, and the absolute-valued difference between this and the corrupted sinogram (darker shade corresponds to higher error). Lower left: true (reference) image, corresponding to the true sinogram. Lower right: reconstruction from the noisy data with FBP algorithm. Images are displayed in HU window $[-220,350]$.}
\label{fig:BK-demo1}
\end{figure}

In the sinogram domain, the standard deviation of the error between the ideal sinogram and the one computed from the measurements, $\hat{g}-\hat{\bar{g}}$, is $\lambda^{-1/2}$ \cite{Maco83}, and this is well approximated by $\hat{y}^{-1/2}$ \cite{LiLi04}. In Figure \ref{fig:BK-demo1} we display a sinogram matrix and the corresponding Poisson noise image. Below, one can observe the resulting reconstruction artifacts produced by the standard FBP algorithm (see next sub-section). The sinogram error image has a high-energy regions where the sinogram values are relatively high; this corresponds to the predicted behavior of the noise variance. The reconstruction from the noisy sinogram is contaminated with anisotropic noise, mainly in the form of streaks. Their appearance is related to large errors in sinogram values.

%\end{document}

%==========================================================================================================================

\subsection{Reconstruction Algorithms for Computed Tomography} \label{sect:BK-recon}

There are various reconstruction algorithms that aim at computing the attenuation map of the scanned object from its projections. In this paper we shall refer and work with two such algorithms: (i) the Filtered Back-Projection (FBP) (\cite{ShKr78}, which is a direct Radon inversion approach. This is a popular technique despite its known flaws; and (ii) an iterative reconstruction algorithm that takes the statistical nature of the unknown and the noise into account (e.g. \cite{ElFe02}). Bayesian methods achieve better image quality than the direct Radon inversion, at the expense of longer processing time. We now describe these two methods is somewhat more details.

\noindent {\bf Filtered-Back-Projection Method:} Mathematically, FBP is the linear operator of the form
\begin{equation}\label{eq:FBP-formula}
\mathbf{T}_{\text{FBP}}=\mathbf{R}^*\mathbf{F}_{low}\mathbf{F}_{RL}.
\end{equation}
Here $\mt{R}^*$ is the adjoint of the Radon transform, known in the literature as "back-projection". $\mathbf{F}_{RL}$ is a 1-D convolution filter, applied to each individual projection (column in the sinogram matrix). It uses the Ram-Lak kernel $k$ \cite{RaLa71}, defined in the Fourier domain by $\hat{k}(\omega)=|\omega|$, and $\mathbf{F}_{low}$ is a low-pass filter which prevents the noise amplification at high frequencies, typical for the Ram-Lak action. In clinical CT scanners, the parameters of $\mathbf{F}_{low}$ are tuned for specific needs of the radiologist: different preset values are chosen to view bones, soft tissues, high contrast/smooth images, specific anatomical regions, etc.

Without the low-pass filter, the FBP is an exact inverse of the Radon transform in the continuous domain \cite{NaWu01} for the noiseless case. Moving from theory to practice, the FBP algorithm does not perform very well. The low-pass 1-D convolution filter in the sinogram domain is not an effective remedy for the projections noise. The problem of photon starvation manifests through outlier values in the sinogram, which propagate to the output image in the form of streak artifacts. They corrupt the image contents and jeopardize its diagnostic value. Those artifacts can be explained as follows: each measured line integral is effectively smeared back over that line through the image by the back-projection; an incorrect measurement results in a (partial) line of wrong intensity in the image. Typically, the streaks radiate from bone regions or metal implants.

\noindent {\bf Statistically-Based Method:} The relation between $f$, the sought CT image, and the vector of measured counts $y$ can be described as
\begin{equation}\label{eq:PL-meas}
\log(y)=\mt{A}f+e,
\end{equation}
where $\mt{A}$ approximates the Radon transform and models the scan process in reality. The additive error $e$ (which also depends of $f$) stems from the statistical noise. In the Bayesian framework, the reconstruction is performed by computing the Maximum a-Posteriory (MAP) estimate of the image
\begin{equation}\label{eq:MAP}
\tilde{f} = \arg\max_f\mt{P}(f|y) =\arg\max_f\dfrac{\mt{P}(y|f)\mt{P}(f)}{\mt{P}(y)}.
\end{equation}
For CT, an accurate statistical model for the data is quite
complicated  and is often replaced by a Gaussian approximation
 with a suitable diagonal weighting term
whose components are inversely proportional to the measurement
variances. This leads to a penalized weighted
least-squares (PWLS) formulation,  see e.g.\cite{RamaniFessler2012}
\begin{equation}\label{eq:PWLS}
\tilde{f} = \arg\min_f \| \log(y) - \mt{A}f \|_D +\beta R(f),
\end{equation}
where  $ \| u\|_D=u^T \mt{D}u$, $\mt{D}$ is a diagonal matrix of weights, which in simplistic model are proportional to photon counts $y$;
The penalty term $R(f)$ also referred to as the \textit{prior}, expresses assumptions on the behavior of the clean CT image. In \cite{Fess06} this expression is chosen as
\begin{equation}\label{eq:BK-penalty}
R(f)= \sum_q\sum_{k\in \mc{N}(q)}\psi_\delta(f_q-f_k),
\end{equation}
where for each image location $q$,
%a scalar function $\psi$ (e.g. $\psi(x)=|x|$) is applied to the local differences to its neighborhood.
a scalar function $\psi_\delta(x)$ is the convex edge-preserving Huber penalty
\begin{equation}\label{eq:FR-huber}
\psi_\delta(x)=\left\{ \begin{array}{ccc}
\dfrac{x^{2}}{2},               &    |x|<\delta \\ \nonumber
\delta|x|-\dfrac{\delta^{2}}{2},&    |x|\geq\delta
\end{array}\right\},
\end{equation}
%Elbakri and Fessler have used in \cite{ElFe02} the method of surrogate functions to minimize \eqref{eq:PWLS};
In order  to minimize \eqref{eq:PWLS}, we have used the L-BFGS optimization method \cite{NoWr06}. The Matlab/C implementation of the algorithm is the courtesy of Mark Schmidt.
%We tuned the parameters $\gamma_H,\delta$ of the PWLS penalty component manually using the set of training images, relying on considerations of the best visual impression .

%==========================================================================================================================
%==========================================================================================================================

\section{Artificial Neural Networks (ANN)}\label{sect:bkgnd-ANN}

For completeness of this paper, we provide here a brief background on ANN, and in particular their role in CT and medical imaging. ANN, mimicking after the biological networks of neurons which comprise the nervous system, are intensively used in many domains of Computer Science. In this work we focus on the multi-layer feed-forward ANN with no cycles. This is best represented by a directed, weighted graph which has an array of input nodes (data inputs), inner nodes (neurons) implementing specific (linear or non-linear) scalar functions, and another array of output nodes. The input argument of each neuron is the weighted sum of all its inputs, where the weights are associated with the edges. Those weights are learned during the network training and, effectively, define the regression function produced by the ANN.

More specifically, the first layer consists of $m$ inputs, coming from the outside world; then $N_l$ neurons are situated in the $l$-th layer ($l>1$), and the last one contains $n$ output nodes. Each input $x_i$ is connected to each neuron $j$ in the second (hidden) layer by a weighted edge with weight $w^1_{i,j}$. The output of each neuron is connected to the input of every $k$-th neuron in the second layer by the weight $w^2_{j,k}$, and so on. Finally, each neuron of the last layer is connected to the output $y_s$ with a weight $v_{s,j}$. We denote by $\sigma$ the function implemented in each neuron. There is a number of popular choices for this function, for instance $\sigma(x)=\tanh(x)$.

\begin{figure}[htbp]
\centering
\includegraphics[width=\columnwidth]{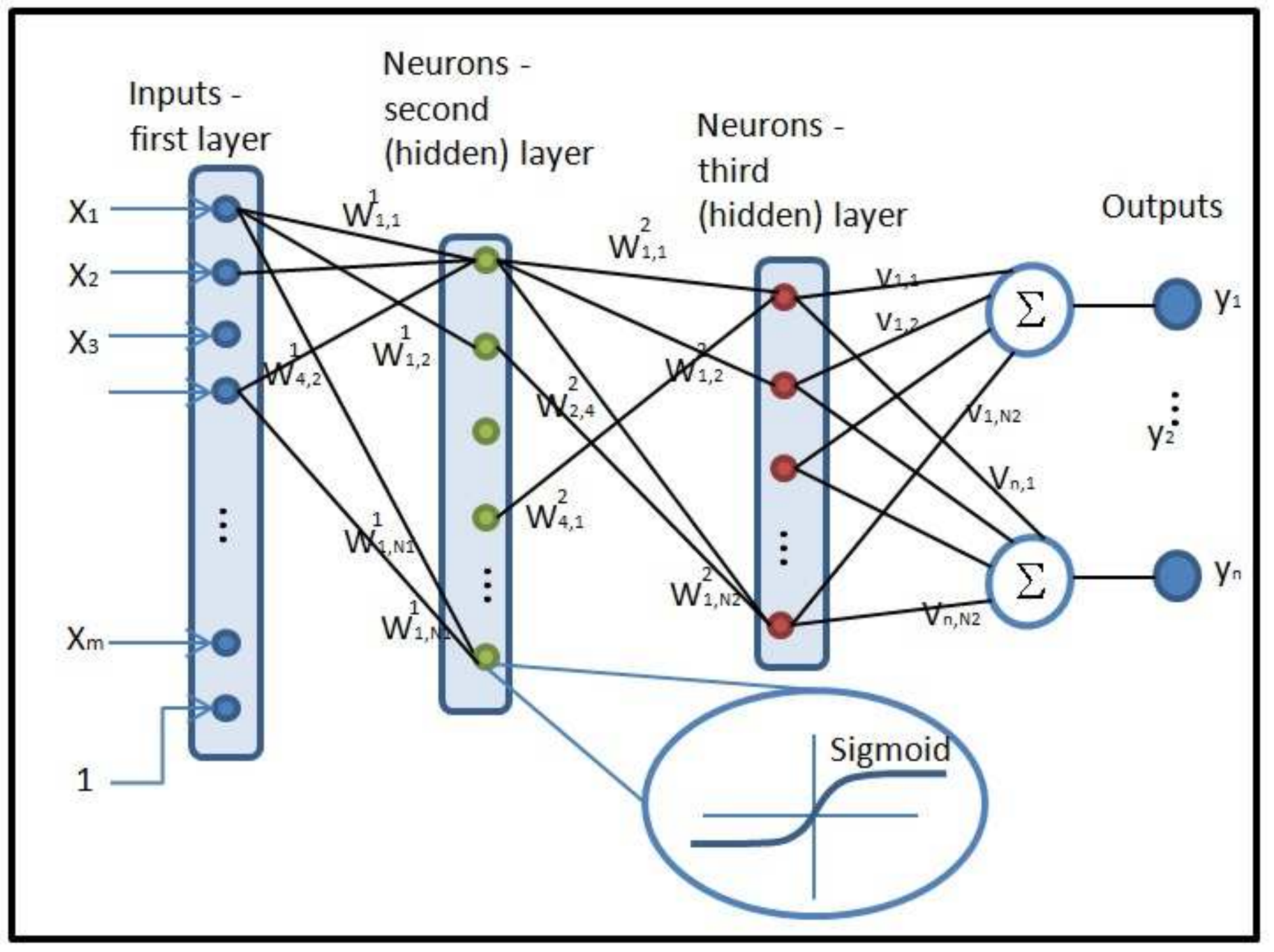}
\caption{A scheme of a multi-layer feed-forward ANN.} \label{fig:ANNsc}
\end{figure}

For example, here is the explicit definition of a network with one hidden layer:
\begin{equation}\label{eq:BK-ANN}
y(x;w,v,b) = \sum_j v_j\sigma\left(\sum_i w_{i,j}x_j+b_j\right).
\end{equation}
%The constants $b_j$ are usually set to $1$.
The weights $\{w,v,b\}$ define the multi-variable regression function $y=y(x)$ which approximates any continuous function implied by the set of training examples\footnote{The Universal Approximation Theorem states that a network with just one hidden layer, where each neuron is realized as a monotonically-increasing continuous function, can uniformly approximate any given multivariate continuous function up to an arbitrary small error bound \cite{Cybe89}. In practice, adding hidden layers shows an improvement in the ANN performance.}. A training set for the network comprises of a collection of examples $(X^k,Y^k)$, where $X^k$ is the vector of inputs and $Y^k$ is the true output related to this vector. Training the network consists of optimizing the weights $\{w,v,b\}$ for a minimal error,
\begin{equation}\label{eq:nn_train}
(w,v,b) = \arg\min_{w,v,b}\sum_{k=1...K}  E\left(y(X^k; w, v,b), Y^k\right),
\end{equation}
where the sum is over the training set, and E(a,b) is an error measure of some sort (e.g. $E(a,b)=(a-b)^2$). The popular method for solution of this problem is the iterative backpropagation method \cite{RuHi86}. A scheme of such network is depicted in Figure \ref{fig:ANNsc}.

Since the development of the back-propagation algorithm for ANN in mid-eighties, the image processing community (among others) has attained a powerful tool to attack virtually any regression or discrimination task. Among the wealth of applications neural networks found in this area (see \cite{EgRi02} for a broad and comprehensive overview), some were designed for medical imaging. As such, Hopfield ANN were used for computer-aided screening for cervical cancer \cite{CeNa00}, breast tumors \cite{BrKa00} and segmentation \cite{ChLi96}. ANN are also used for compression and classification in cardiac studies \cite{HiMa94} and ECG beat recognition \cite{OsHo01}. Tasks of filtering, segmentation and edge detection in medical images are addressed with cellular ANN in \cite{AiAi01}. Our group has used neural networks for optimal photon detection in scintillation crystals in PET  \cite{nnpet}.

As for reconstruction problems, a series of works has appeared in which the ANN replaces the overall reconstruction chain by learning the net contribution of all detector readings to each pixel in the image. For Electron Magnetic Resonance (EMR), such an algorithm is proposed in \cite{DuKr07}. Floyd et. al. have used this approach for SPECT reconstruction \cite{Floy91} with feed-forward networks and also for lesion detection in this modality \cite{FlTo92}. We remark that such naive application of the ANN for reconstruction is limited to low-resolution $n\times n$ images, since the network must have $\mc{O}(n^2)$ inputs and outputs. For instance, in \cite{DuKr07}, a $64\times 64$ image is reconstructed. Application of ANN for SPECT reconstruction was also studied by J. P. Kerr and E. B. Bartlett \cite{KeBa95a,KeBa95b}.

Imaging modalities like PET and SPECT, where low-resolution images are produced, are a natural domain for ANN application. However, some works tackle also the problem of CT reconstruction where the image size is larger. Ref. \cite{Cier08} proposes using a neural network structure with training based on a minimization of a maximum entropy energy function. Reconstruction in Electrical Impedance Tomography was treated with ANN in \cite{AdGu94}. Another variety, an Electrical Capacitance Tomography and an ANN-based reconstruction method for it, are described in \cite{NoHo97}.

Despite the abundance of applications, there is still place for innovation in the domain of ANN application for medical imaging. First, the CT reconstruction problem is rarely attacked with this tool due to the high dimensions of raw data and the resulting images, which render the naive application of ANN as the black box converting measurements to image unfeasible. Indeed, in our work we do not propose such a scheme per se -- rather, our ANN is employed to perform a locally-adaptive fusion of a number of image versions, produced by a given reconstruction algorithm upon using different configurations. This brings us naturally to the next section where we describe our algorithm.

%==========================================================================================================================
%==========================================================================================================================

\section{The Proposed Scheme}\label{sect:RVTF-concept}

%==========================================================================================================================

\subsection{Local Fusion with a Regression Function}

We consider the general setup of the non-linear inverse problem. Assume we are
given the measurements vector $y$ of the form
\begin{equation}\label{eq:RV-problem}
y = \mt{H}x+\xi,
\end{equation}
where $\mt{H}$ is some transformation, $\xi$ represents the noise, and $x$ is the signal to be recovered. Assume further that $\mt{T}_{\mt{p}}$ is some restoration algorithm designed to recover $x$ from this type of measurements, i.e.,
\begin{equation}
\bar{x}_{\mt{p}} = \mt{T}_{\mt{p}}(y)
\end{equation}
The scalar parameter $\mt{p}$ controls the behavior of $\mt{T}$ and therefore influences certain characteristics of the estimate $\bar{x}$. For example, when $\mt{p}$ is responsible for variance-resolution tradeoff of the algorithm, the estimate $\bar{x}_{\mt{p}}$ may be obtained with different noise levels and corresponding spatial resolution characteristics.

The described situation is common in many signal/image processing scenarios. As a basic example, we consider a simple image denoising algorithm, which recovers the signal $x$ from noisy measurements $y= x+\xi$ by a shift-invariant low-pass filter, realized as a 2-D convolution with prescribed kernel. For some fixed shape of this kernel (say, a simple boxcar function or a 2-D Gaussian rotation-invariant kernel), its width (spread) can be parameterized by a scalar variable $\mt{p}$. A wider such kernel will perform a more aggressive noise reduction, by averaging the noisy signal over a larger area, at the cost of reducing the spatial resolution.

A second, and more relevant example to this work, is from the domain of CT reconstruction. Recovery of the attenuation map is classically performed by the Filtered Back-Projection algorithm. The latter involves a 1-D low-pass filter, applied to the individual projections. As in the above example, the cut-off frequency of this filter controls the variance-resolution properties of the reconstructed image. In these examples, and also in a general such situation, no single value for the parameter $\mt{p}$ makes the best of the processing algorithm. For different signals, different values may be
optimal in the sense of MSE or other quality measure. Indeed, in the same image, computed with two different values of $\mt{p}$, different regions will get the best treatment by different values of $\mt{p}$. For each specific case, ad-hoc considerations for tuning this scalar parameter are applied.

In the domain of non-parametric statistics, there is a noise reduction algorithm with proven near-optimality that devises a switch rule for selecting at each location of the signal an appropriate local filter \cite{GoNe97}. In effect, the signal is processed by a low-pass filter adaptive to the local signal smoothness. In the context of our discussion, one can say that this algorithm performs a fusion of a number of filtered versions of a signal with varying filter parameter. The switch rule, developed for this adaptive signal smoothing, is based on the balance of the stochastic and structural noise components and model assumptions, and as such, it is very difficult to devise. Moreover, better output may be obtained if we allow to use some combination of the given image versions in each pixel, rather than selecting one of them alone. To our knowledge, no mathematical theory offering a descriptive rule for such local fusion is available for signal estimators, used for denoising or CT reconstruction.

Borrowing from the above switch-rule idea between filters, the solution we propose for the problem described above is a local fusion of a sequence of estimates $\bar{x}_{\mt{p}_1},...,\bar{x}_{\mt{p}_N}$ with a specific regression function, learned on a training dataset consisting of similar cases. Among known regression methods, we choose to work with ANN, due to their strong adaptivity and generalization properties \cite{Hayk94}. The supervised learning is done with a training set of examples: For each location in the processed signals, the features (input vector) are sample values extracted from the corresponding location in the sequence of reconstructed versions for this signal. The output is a small region of sample in the desired destination signal. Contemporary training algorithms employ error back-propagation to optimize the objective function, measuring the discrepancy between the correct output values and those predicted by the ANN \cite{RuHi86}. In our work we employ the Matlab Neural Network toolbox; the training was performed with the Levenberg-Marquardt  algorithm \cite{Marq63,Gavi11}. Our networks consist of two hidden layers. We use the function $\sigma(x)=x/(1+|x|)$, which has similar properties to the classical sigmoid and is computationally cheaper and is more robust to saturation caused by large arguments.

In this work, the outlined general concept is specialized to reconstruction algorithms for CT. Specifically, we consider representatives of the two types of those algorithms: the direct FBP and the iterative PWLS (Section \ref{sect:BK-recon}) methods. For FBP, we propose making a sweep over the cut-off frequency of its low-pass filter in the sinogram  domain. This parameter controls the noise-resolution tradeoff and has a major influence on the visual impression of the resulting images. For the iterative PWLS algorithm, a sequence of images is extracted along its execution by saving a version of the CT result every few iterations. In following sections we illustrate this approach on a simple 1-D denoising problem and work out a number of applications for CT reconstruction algorithms, as detailed above. Along the way, we discuss the choice of training set and design of features extracted for the ANN.

%==========================================================================================================================

\subsection{An Example: ANN Fusion for 1-D Signal Denoising}\label{sect:RVTF-PWC}

To illustrate the proposed concept, we begin with the simple signal denoising algorithm as mentioned above. We assume that the original signal is 1-D piece-wise constant (PWC). This choice is beneficial for the test we are about to present, since random PWC signals can easily be generated for training/testing purposes, and the effect of low-pass filter denoising is easily observed. We generate such a signal $x$ of length $n$ by choosing $n/30$ step locations uniformly in random, and choosing the intensity value for each step uniformly at random as well, in the [0,1] segment.

\begin{figure}[htbp]
\centering
\includegraphics[width=1\columnwidth]{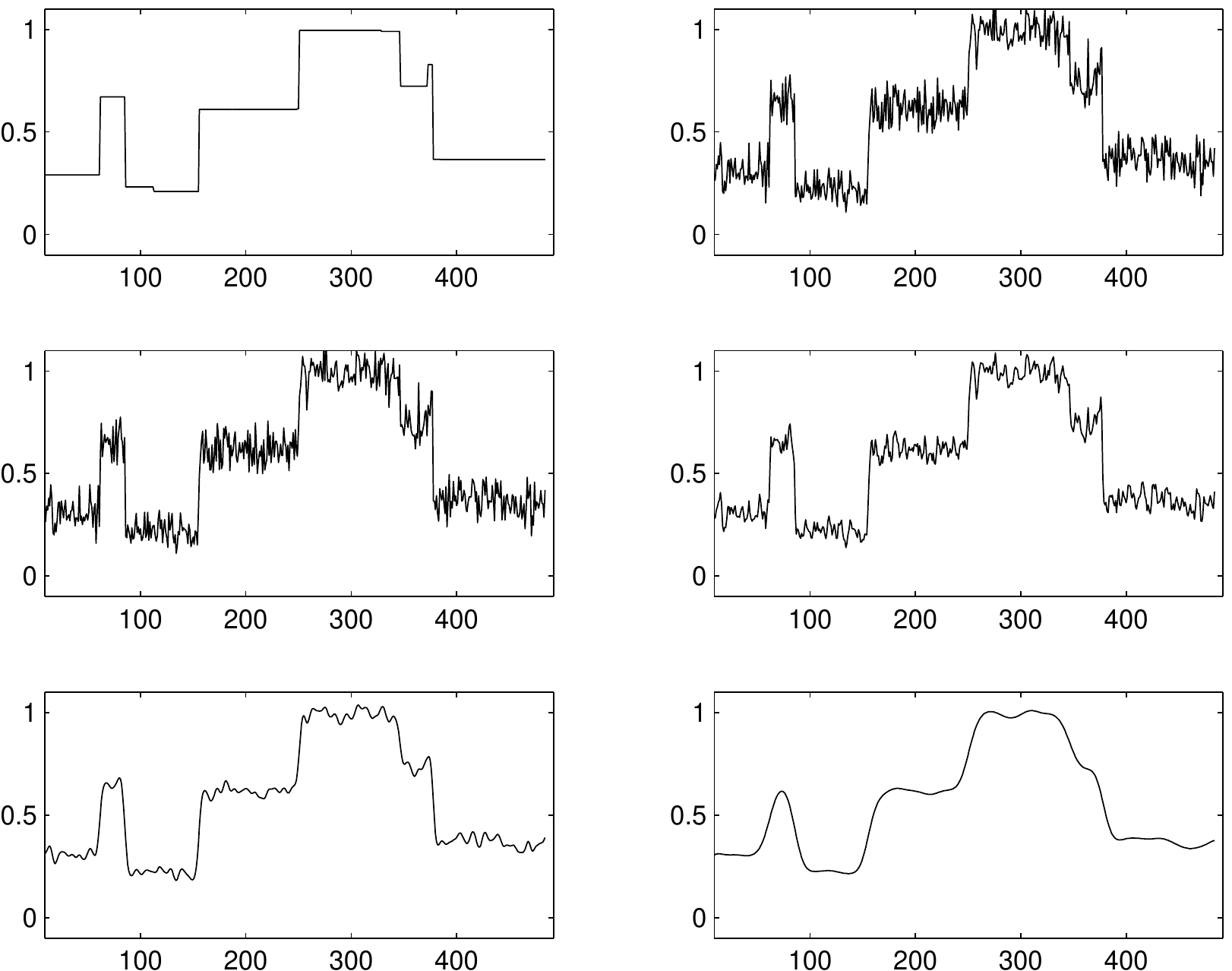}
\caption{Left to right, top to bottom: clean train signal $x$, noisy signal $y$, and several restored signals $\hat{x}_i$ obtained by convolution with a Gaussian kernel of increasing width.} \label{fig:RVTF-PWC-1}
\end{figure}

Assume that such a signal $x$ has been created and is contaminated with i.i.d. Gaussian noise $\xi \sim\mc{N}(0,\sigma_n)$ with $\sigma_n=0.06$. For the noise reduction, we perform a convolution of $y=x+\xi$ with a Gaussian kernel $G(p)=\mc{N}(0,p)$. For some chosen values of the standard deviation $p=p_1,...,p_{8}$ we obtain the sequence of estimates
\begin{equation}
\hat{x}_i = y*G(p_i),\;\;\;i=1,...,8.
\end{equation}
In Figure \ref{fig:RVTF-PWC-1} we display an instance of such a signal, the corresponding noisy version, and a number of signal estimates obtained with convolution filters of different widths.

In this setup we train the ANN for a better signal restoration. For each location $q$ in $y$, we extract a set of small neighborhoods of the same location $q$ from each of the signals $\hat{x}_1,...,\hat{x}_K$. Those are concatenated into one vector which serves as the ANN input. Specifically, we take a $11$-samples window from each processed signal in the sequence of $K=8$ signals. Thus, overall the feature vector for each location is of size $8\cdot11=88$ samples. In the training stage, every such vector is matched with a label -- the correct value $x(q)$, which is provided to the ANN as the desired output.

\begin{figure}[htbp]
\centering
\includegraphics[width=0.8\columnwidth]{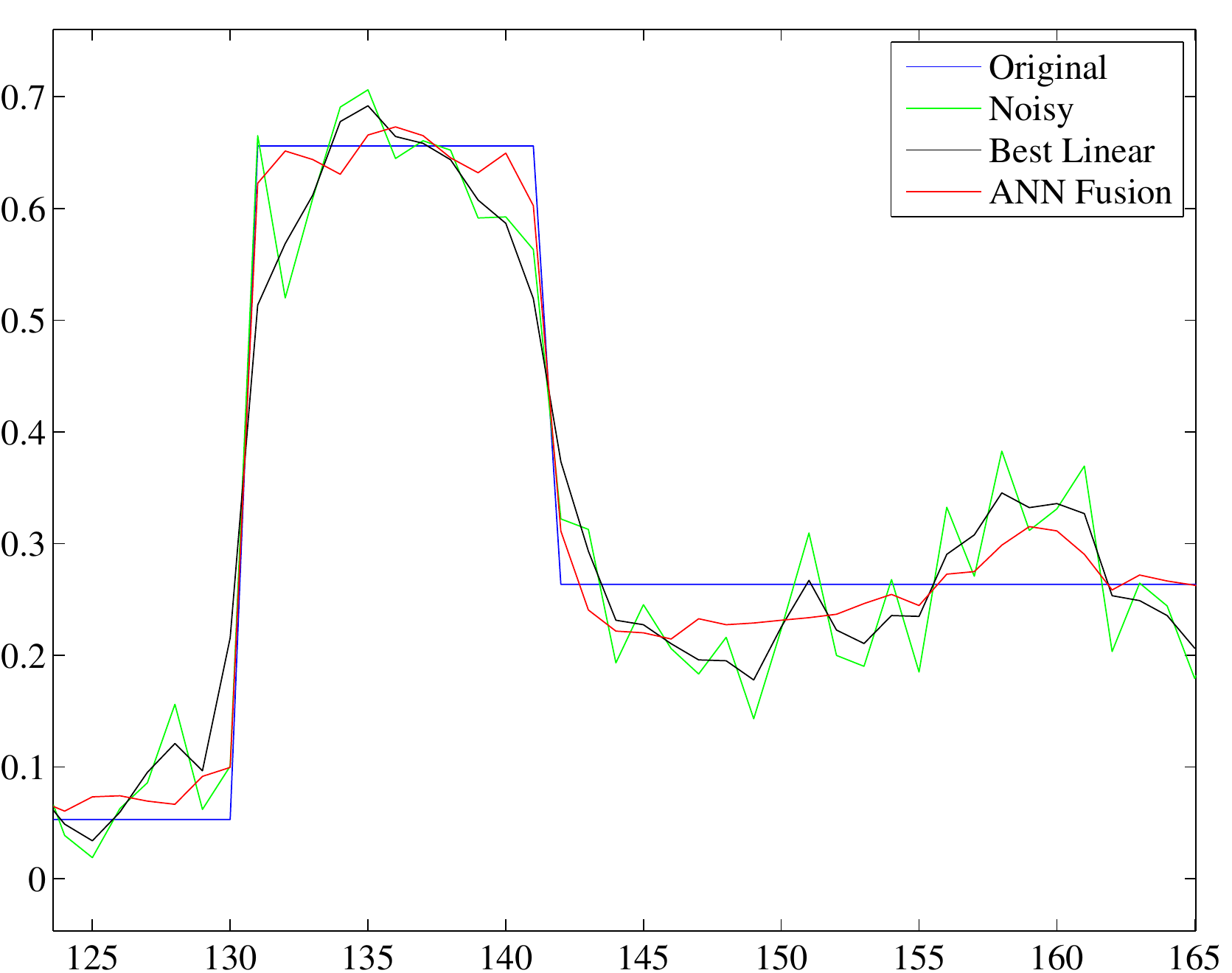}
\caption{Test results: a zoom-in on a portion of the clean signal $x$, the noisy one $y$ ($17.67$dB), the SNR-best linearly restored signal
$\hat{x}_{i}$ ($19.85$ dB), and the restoration by the ANN ($26.18$dB).}
\label{fig:RVTF-PWC-3}
\end{figure}

For the training procedure we generate a signal $x$ of size $n=2\cdot 10^4$ (=number of training samples) and extract the training data as described above.  The obtained ANN is tested on another signal of length $300$, randomly generated with the same engine. In Figure \ref{fig:RVTF-PWC-3} such test results are presented. The neural network has improved the SNR of the best linear estimate from $19.85$dB to $26.18$dB, and this difference is observed in the fact that the ANN estimate fits the original signal much closer. The SNR values are calculated over an interval of $200$ samples in the center of the test signal, so as to avoid boundary problems.

The presented algorithm has various design variables: the number, shape and width of the applied filters, the size and structure of the neural network, the structure of a input vector for each example (set of features). The questions of algorithm design will be pursued in the following sections, where CT reconstruction algorithms, relevant to our study, are invoked in the similar setup of
performance boosting by local ANN fusion.

%==========================================================================================================================

\subsection{Error Measures}

Just before we conclude this section and move to present the specific details of boosting CT reconstruction algorithms, we should discuss the choice of the error function to use in the learning process, and the error measure to use when evaluating the quality of the reconstruction.

\vspace{0.1in}
\noindent {\bf C.2 Training Risk}

In the supervised learning procedure, we design the ANN weights so as to minimize the regression error between the ANN output and the desired labels (training output data). In many cases, the natural choice for this function would be the Mean-Squared-Error (MSE). However, in CT, we should contemplate whether MSE is the proper choice to use. Consider a homogeneous region in a CT image (corresponding to some tissue) with a small detail of a different yet similar intensity (a cavity or a lesion). The MSE penalty paid by an over-smoothing reconstruction filter that blurs this lesion is small, and therefore such faint details may be lost while leading to better MSE. The remedy for this problem could be to penalize not only for the difference in intensity values between the reference image $f_0$ and the reconstruction $\tilde{f}$ , but also for the difference in the derivatives of these two images. Alternatively, We can weight the training examples so as to boost the importance of such faint edge regions, at the expense of more pronounced parts of the image, where the edges are sufficiently strong. In this spirit, building on the general error term written in Equation (\ref{eq:nn_train}), we propose to use
\begin{eqnarray}\label{eq:grad-penalty}
\theta^* & = & \arg\min_{\theta}\sum_{k=1...K}  E\left(y(X^k,\theta),Y^k\right) \\
& = & \arg\min_{\theta}\sum_{k=1...K} \rho_k \cdot \left(y(X^k,\theta)-Y^k\right)^2. \nonumber
\end{eqnarray}
In this expression $(X^k,Y^k)_k$ is the training data consisting of pairs of feature (input) vectors and their desired label (output), and the function $y(X^k,\theta)$ is the output of the ANN, governed by its control parameters $\theta$. This is a simple weighted MSE, and the idea mentioned above is encompassed in the choice of $\rho_k$, the scalar weights assigned to the training examples. In our work we have chosen $\rho_k$ to be zero for examples having a very low variance in the input image,  which correspond to air regions. Specifically, the threshold is set to $10^{-6}$ times the maximal variance of $X^k$. A zero weight is also assigned to all the examples where the accumulated gradient over the input patch (in the idea image) is above $2\%$ of its maximal value. The later pruning is introduced in order to avoid the bias of the very strong bone-flesh, flesh-air edges in the training process. As for the remaining examples, we assign their weight to be proportional to the accumulated gradient of the patch (again, in the ideal image). This way, the remaining informative edges get a more pronounced effect in the learning procedure.

% ====> There is a problem with this description - it is not accurate at all, and we need to resolve this.

%\noindent {\bf PROBLEM WITH THIS DESCRIPTION (SEE COMMENT IN TEX FILE)}

\vspace{0.1in}
\noindent {\bf C.2 Quality Assessment}

The quality measures of CT images used in this study, are the following:
\begin{itemize}
\item
\textbf{Signal-to-Noise Ratio} (SNR), defined for the ideal signal $f$ and a deteriorated version $\hat{f}$ by SNR$(f,\hat{f})=-20 log_{10}(\|f-\hat{f}\|_2/\|f\|_2)$. In practice, we consider the signal $\hat{f}$ up to a multiplicative constant and compute
\begin{equation}
SNR(f,\hat{f})=\max_{\alpha}-20
log_{10}(\|f-\alpha\hat{f}\|_2/\|f\|_2).
\end{equation}
To make the error measurement more meaningful, the SNR is only computed in the image region where the screened object resides, ignoring the background area. We have used an active contour technique to find the object region in the image; specifically we have used the Chan-Vese method \cite{ChVe01}.

\item
\textbf{Windowed Signal-to Noise Ratio}. The dynamic range of the HU values in a CT image is very large, from $-1000$ for air to $1500-2000$ for bones. Often, the diagnostic interest lies in the soft tissues, the HU values of which are near zero (HU of water). For axial sections of thighs, we chose (by a criterion of best visibility ) the window of $[b_1=-220, b_2=350]$ HU; our algorithms are tuned for best reconstruction in this HU range. Therefore, an appropriate SNR measurement considers only the regions in the image that fall in this range. Technically, the reference image $f$ and the noisy image $\hat{f}$ are pre-processed before the standard SNR is computed by projecting values lower or higher than $b_1$ and $b_2$ respectively to these values.

\item
\textbf{Structured Similarity} (SSIM) measure \cite{WaBo04}. This measure of similarity between two images comes to replace the standard Mean Squared Error  (the expression $\|f-\hat{f}\|_2$ appearing in the SNR formula), which is known for its problamatic correlation with the human visual perception system (see \cite{WaBo04} and the references 1-9 therein). SSIM compares small corresponding patches in the two images, after a normalization of the intensity and contrast. The explicit formula involves first and second moments of the local image statistics and the correlation between the two compared images. In our numerical experiments, we use the Matlab code provided by the authors of \cite{WaBo04}, which is available at \text{https://ece.uwaterloo.ca/}$\sim$\text{z70wang}
\text{/research/ssim}.

\item
\textbf{Spatial resolution measure}: the spatial resolution properties of a non-shift-invariant reconstruction method should be characterized using a local impulse response (LIR) function, which replaces the standard point-spread function \cite{FeRo96}. We compute the LIRs by placing sharp impulses (single pixel) in many random locations in the reference image and by taking the difference between the reconstructed images, scanned with or without the spikes. For each LIR, the Full-Width Half-Maximum (FWHM) value is computed as follows: first, the 2-D image matrix of the response function is resized into an image larger by $\times 16$ in each axis, in order to reduce the discretization effect. Then, the number of pixels with intensity higher than half-maximum is counted and divided by the refinement factor of $256$. This is the FWHM resolution measure at the specific location.
\end{itemize}

%==========================================================================================================================
%==========================================================================================================================

\section{FBP Boost -- Algorithm Design}\label{sect:RVTF-FBP}

%==========================================================================================================================

\subsection{The Low-Pass FBP Filter Parameters}

The method of local fusion, advocated in the previous section, is now applied to the standard Filtered Back-Projection (FBP) algorithm for CT reconstruction. The fusion is performed over the parameters of the low-pass sinogram filter, applied before the Back-Projection. This one-dimensional low-pass filter is realized as a multiplication with the Butterworth window $H$ in the Fourier domain, defined by
\begin{equation}\label{eq:SP-Butterworth}
|\hat{H}(\omega)| = \left(1+\left(\dfrac{\omega}{\phi_0}\right)^{2p}\right)^{-1/2}.
\end{equation}
We sweep through the range of the parameter $\phi_0$ (expressing the cut-off frequency of the filter), thus changing the resolution-variance tradeoff of the FBP. We also change the parameter $p$, which controls the steepness of the window roll-off. While $\phi_0$ controls the amount of blur introduced during the reconstruction, the parameter $p$ influences the texture of reconstructed image.

\begin{figure}[htbp]
\centering
\includegraphics[width=1\columnwidth]{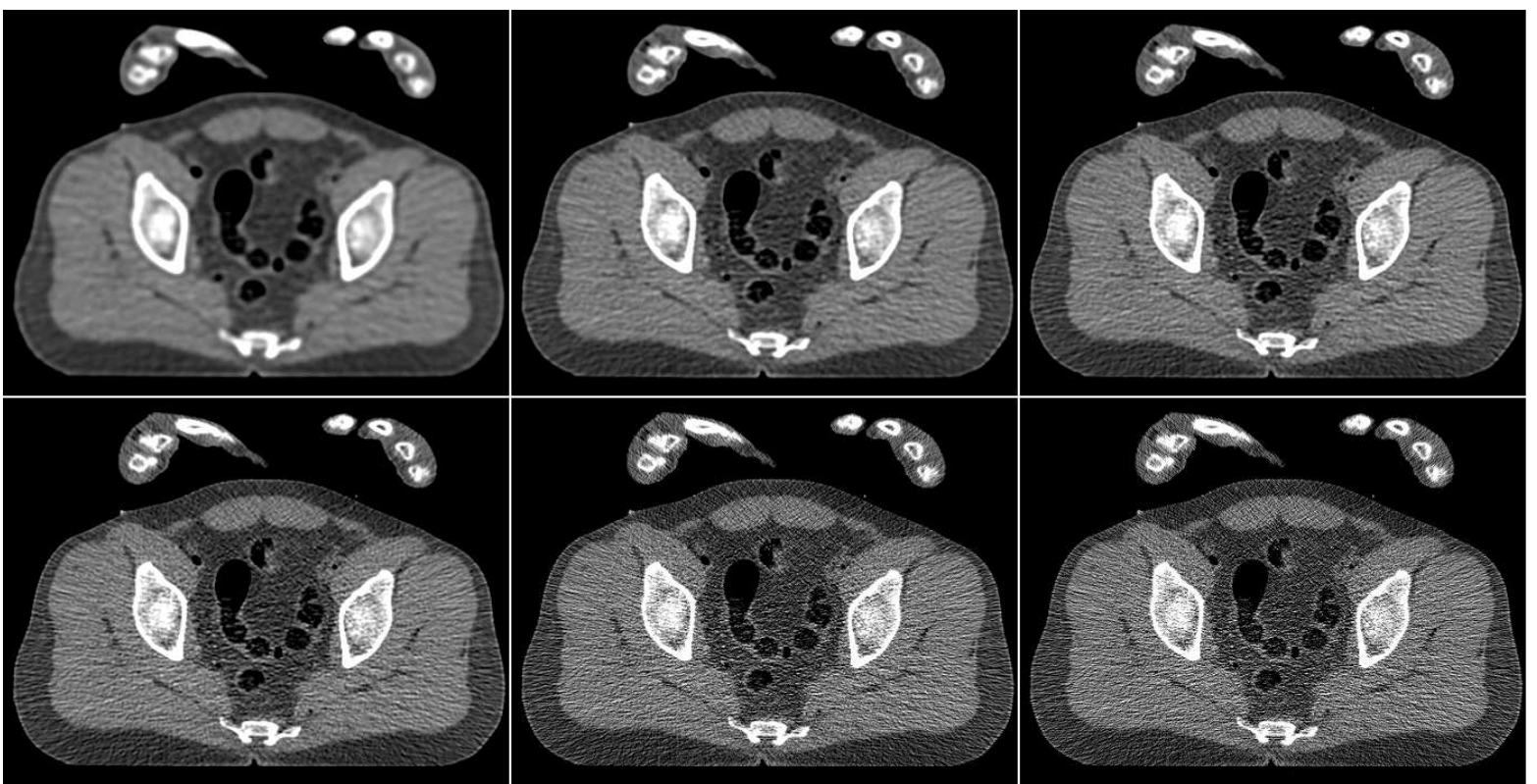}
\caption{FBP reconstruction with different cut-off frequency value. Upper to lower, Left to right: $\phi_0= [0.4, 0.8, 1.15, 2.0, 120,\infty]$ (the last image is compute without the low-pass filter).} \label{fig:RV-FBP1b}
\end{figure}

In Figure \ref{fig:RV-FBP1b} we show the reconstruction for a fixed value of $p=3$ and an increasing cut-off frequency $\phi_0$. Visually, the strong low-pass filter produces a cleaner image (which also have a higher SNR), but looses in the spatial resolution. The displayed sequence corresponds to values $\phi_0=[0.4, 0.8, 1.15, 2.0, 120 ,\infty]$ (the last corresponds to no filter).

After testing various combinations, we chose to use only three FBP images with cut-off frequencies $\phi_0= [0.4, 1.15, \infty]$ and $p=3$. Those were selected from eight images -- three with the frequencies $\phi_0= [0.4, 0.8, 1.15]$ and $p=1$, another three images with the same frequencies and $p=3$, and the last two are obtained with $\phi_0= [2.0, 120]$ and $p=3$. The reason for the restriction to three images is the smaller ANN required.

%==========================================================================================================================

\subsection{Design of The ANN Fusion and Training Setup}\label{sect:SP-ANN-design}

Let $\tilde{f}_1,..., \tilde{f}_K$ be a given set of versions of a CT image, reconstructed by FBP with different low-pass filters in the sinogram domain.\footnote{Note that all these images are produced from the very same raw sinogram, which means that the patient is exposed to radiation only once.} We describe the fusion procedure used to compute the output image $\hat{f}$ of the algorithm:
\begin{itemize}
\item
For each location $q$ in the image matrix, extract its disk-shaped neighborhood from each of the $K$ images $\tilde{f}_i$, $i=1,...,K$. The radius of the disk is set to $3$ pixels (containing $29$ pixels).
\item
Compose a set of inputs for the ANN by stacking the pixel intensities from the $K$ neighborhoods into one vector. Normalize this vector in the training stage (discussed below).
\item
Apply the ANN to produce a set of output values, which are the intensity values in the disk-shaped neighborhood of $q$ in the image $\hat{f}$. This disk has the same radius of $3$ pixels.
\item
By this design, each pixel in the output image is covered by $29$ disk-shaped patches; its final value is computed by averaging all those contributions.
\end{itemize}

We detail now on the several of the steps in the list above. In the training stage, the neural network is tuned to minimize the discrepancy between true values in each output vector and those produced by the network from the set of noisy inputs. A vector of inputs is built, as described above, for a location $q$ in a reference image $f$ from a training set, using data from noisy reconstructions. The corresponding vector of outputs is the disk-shaped neighborhood of $q$ in the reference image. Thus, for each image $f$ we produce the set $\tilde{f}_1, ...,\tilde{f}_K$ using pre-defined FBP filters and sample them to build the training dataset. The image is sampled on a cartesian grid, choosing every third pixel $q$ both in horizontal and vertical directions. The pair of input and output vectors for the neural networks is an example used in the training process. Examples from all the training images $f$ are put in one pool. A portion of this pool, having a very low variance in the inputs vector, is discarded (specifically, the threshold is set to $10^{-6}$ times the maximal variance). Those examples correspond to regions of air, since no constant patch in any kind of tissue can be observed in the noisy FBP images. This step leads to an empirical improvement in the performance of the ANN.

It is generally acknowledged, that data normalization improves performance of neural networks \cite{lecun1998efficient}. Our data matrix $A$, which columns are the individual example vectors, is  normalized by
\begin{equation}
A \Leftarrow A - \min_i(A(i))\mbox{  and then   } A \Leftarrow A/\max_i(A(i)).
\end{equation}
The two constants $\alpha_1 = \min_i(A(i))$ (the minimum value of the matrix $A$) and $\alpha_2 = 1/\max_i(A(i))$ are stored along with the weights of the neural network, and the new data matrix in the test stage is transformed with those pre-computed constants.
%Such processing enables significant improvement of the neural network's performance.

Given intensity values in the neighborhood of a pixel $q$ in several noisy images, the network should predict a single value in this pixel for the fusion image. However, as a step of regularization, we design the ANN to produce a vector output which is interpreted as a small neighborhood of $q$. the fusion image is then built from such disk-shaped overlapping patches, which are averaged to produce the final result. This is done to avoid possible artifacts, which can be produced by the network: in the training stage, if the ANN produces a single outlier intensity value, its penalty will be smaller than of a vector of such incorrect intensities. Such regularization reduces the performance the ANN can achieve on the training set, since more equations are imposed, but its performance on test images is expected to be more stable.

%==========================================================================================================================
%==========================================================================================================================

\section{FBP Boost -- Empirical Study}

\subsection{Evaluating the Algorithm Performance}

In the experiments we have used sets of clinical CT images, axial body slices extracted from a 3D CT scan of a male head, abdomen and thighs. The images are courtesy of Visible Human Project. The intensity levels of those grayscale images correspond to Hounsfield Units. The training set comprises of $461\times 461$ male thighs sections. The image set for ANN training consists of $12$ images, from which $30,000$ examples are extracted. This number, in our experience, suffices to avoid an over-fitting for the chosen size of neural network ($40$ neurons in
the hidden layer, $90$ network inputs, overall $3720$ weights). The vector of features for each example is built from the pixel neighborhoods of radius $3$ pixels, coming from the three corresponding FBP reconstructions. These images are a subset of the $8$ FBP reconstruction images mentioned before, seeking (manually) for the subgroup that would perform the best. The size of the input vector is $3\times 29 = 87$ entries.

\begin{figure}
\centering
\includegraphics[width=1\columnwidth]{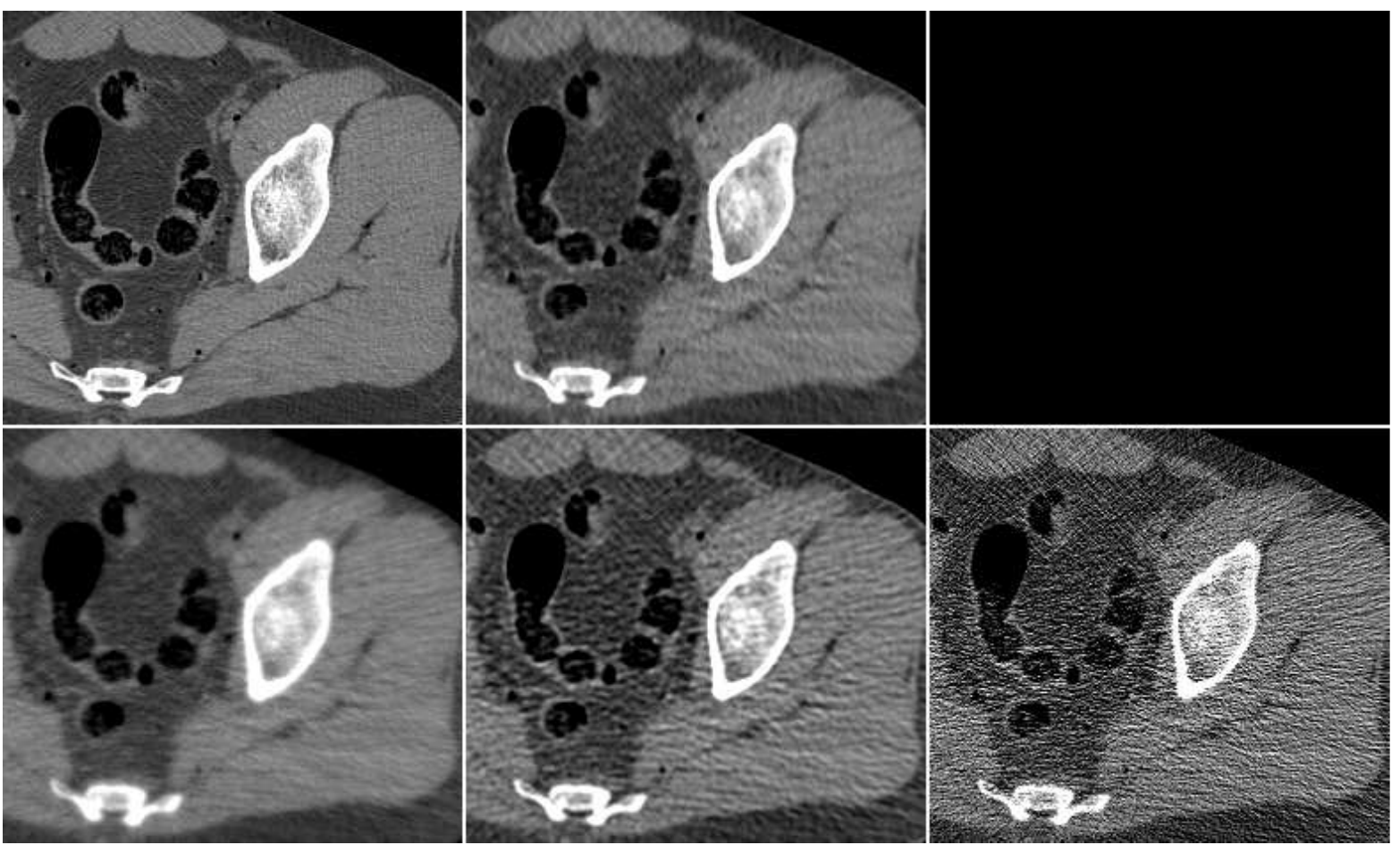}
\caption{Upper left: reference image. Upper middle:  the ANN fusion result. Other: FBP images participating in the fusion, produced with different low-pass filters.} \label{fig:SPF1}
\end{figure}

In Figure \ref{fig:SPF1} we present a reconstruction of a test image. This test image is taken $10$cm away from the region where the training data was taken from. The middle upper image is the result of a fusion of the number of FBP versions, performed with the trained ANN. By the visual impression, the noise-resolution balance in the fused image $\hat{f}$ is better than in any of the FBP versions forming it. The texture of tissues is closer to the original (observed in the reference image, upper left). The level of streaks and general noise are lower than in the central and right FBP images, and the image sharpness is higher than in the left and the central images. Thus, the fusion image enjoys the good properties offered by each of the FBP versions and is superior than any of them.

Recall that the training was done with a set of weights, corresponding to our penalty component from Equation \ref{eq:grad-penalty}. The quantitative error measures we compute for this comparison include plain SNR, SNR weighted by those weights, the training risk and the SSIM measures. These values are given in Table \ref{tbl:SPF-parade1}. As observed from the table, the weighted SNR of the fusion image is $1.8$dB higher than the highest attainable value in FBP images. For plain SNR this increment is $1.5$dB. Values of the training risk measure behave expectedly: the weights of ANN training were designed to implicitly reduce this measure for the fusion image. Indeed, it is by $20\%$ lower than that of the optimal FBP image. Finally, the SSIM measure supports the claim the fusion image has the best visual appearance, since it admits the larger value for this measure.

\begin{table}[htbp]
\begin{center}
\begin{tabular}{||l|l|l|l|l||}
\hline \hline
Image  & FBP & FBP & FBP & Fusion  \\
 & $\phi_0=0.4$ & $\phi_0=1.15$ & $\phi_0=\infty$  &  result \\
\hline \hline
SNR (uniform)  &  25.3059 &  22.1515  & 19.2833 &  26.8692\\
\hline
SNR (weighted) &  24.3437 & 22.3835 &  19.4414  & 26.1060\\
\hline
Training-Risk  & 40.6049 & 25.3577  &  55.7256 &  20.5624\\
\hline
SSIM  &  0.8839 &  0.8939  &  0.6892  &  0.9298\\
 \hline  \hline
\end{tabular}
\end{center}\caption{Quantitative measures for the FBP reconstructions and the fusion result.}
\label{tbl:SPF-parade1}
\end{table}

%==========================================================================================================================

\subsection{Size of Local Neighborhood}

We study the algorithm performance with different amounts of local data provided for the ANN fusion. A sequence of test image reconstructions is produced, where the radius $r$ of the pixel neighborhood, extracted for the fusion, is increased from $r=0$ (single pixel) to $r=4$ ($49$ pixels). The input vector for the ANN is built from three such neighborhoods, coming from FBP reconstructions corresponding to cut-off frequencies $\phi_0 = [0.4, 1.15, \infty]$ of the low-pass filter. We remark that in the special case of $r=0$, the regression function learned by the network incorporates only the relations between the pixel values in the different image versions, while with larger neighborhood sizes there is also a possibility to perform some local filtering in each image.

\begin{figure}
\centering
\begin{tabular}{ll}
\includegraphics[width=0.75\columnwidth]{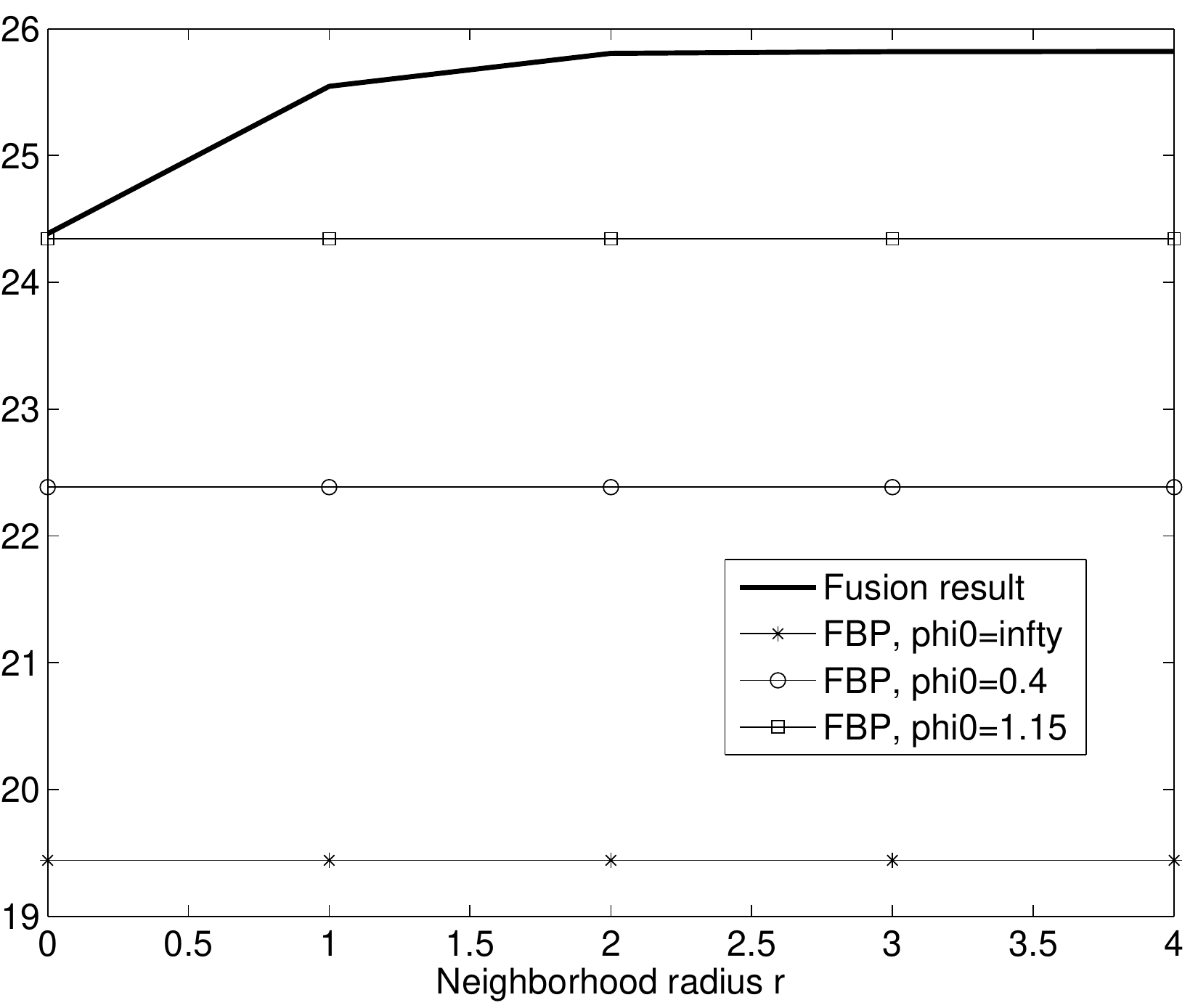}
\end{tabular}
\caption{Graphs of the SNR %(left) and SSIM (right)
values corresponding to reconstructions with ANN fusion using input pixel neighborhoods of radius $r=0,1,2,3,4$ (x-axis).
} \label{fig:SP-r1-seq}
\end{figure}

In Figure \ref{fig:SP-r1-seq} we display graphs of SNR values\footnote{Very similar effect was observed with SSIM.} computed for the test image. Observably, the quality increment with the neighborhood radius is exhausted around $r=4$. Our choice is to use $r=3$, which requires a smaller number of variables (comparing to $r=4$) without almost no loss in quality. We also notice in these graphs that the fusion using only the central pixel $p$ has a performance very close to that of the best FBP version (but slightly higher, which testifies to the necessity to provide a larger neighborhood of each pixel for a successful fusion. We should note that large neighborhood allows the network to perform a kind of directionally anisotropic filtering matched to the direction of edges.

\begin{figure}
\centering
\includegraphics[width=1\columnwidth]{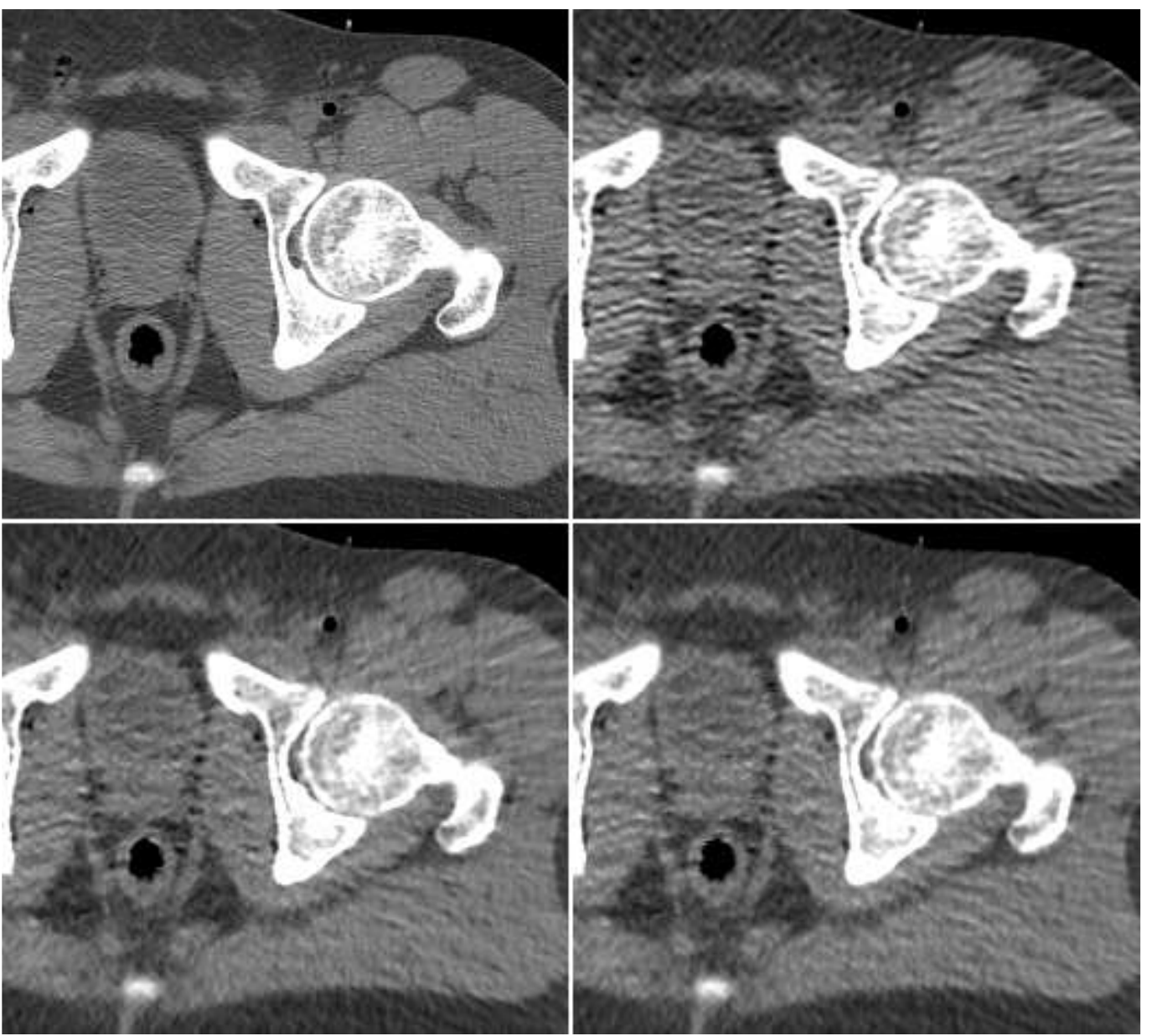}
\caption{Upper left: reference image. Upper right: best-SNR FBP reconstruction. Lower left: fusion result where ANN output size is $5$ pixel. Lower right: a fusion result where the ANN produces a single pixel value.} \label{fig:SP-ym1}
\end{figure}

We also compare two cases of output vectors produced by the ANN. In the lower row of Figure \ref{fig:SP-ym1}, the image on the right is produced by the fusion process where a single pixel is recovered by the ANN for each input vector. The image on the left is produced by computing $5$-pixel neighborhoods of each pixel and averaging the overlapping regions. The visual difference between the image is negligible, and the difference in SNR is $0.2$ dB in favor of the averaging approach. Judging from this (and other similar) tests, we conclude that forcing the neural networks to evaluate a number of pixels in the neighborhood of the one being recovered does not reduce its performance. We don't have an empirical evidence that such a step is truly necessary, since no artifacts in single-pixel-estimation case were observed in this test.

%==========================================================================================================================

\subsection{Single-Image ``Fusion''}

\begin{figure}
\includegraphics[width=1\columnwidth]{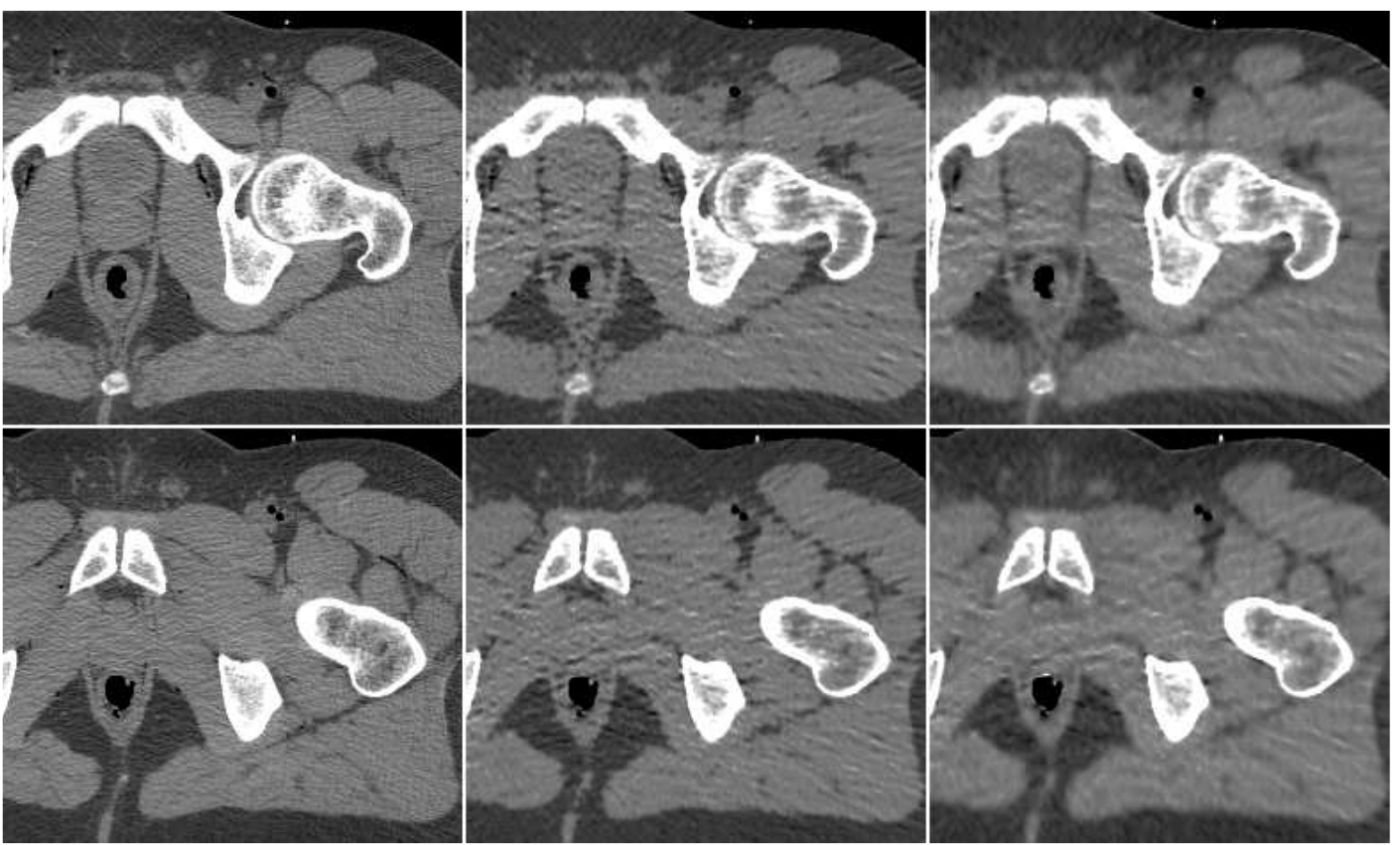}
\caption{Two test images (corresponding to the upper and lower rows) of thighs sections. Left column: reference images. Middle column: ANN fusion of a single FBP image with no sinogram filter ($\phi_0 = \infty$). Quality of the upper middle image: SNR = 26.13 dB; lower middle: SNR = 27.02 dB.
Right column: ANN fusion of three FBP versions, corresponding to filter cut-off frequency of $\phi_0 = [0.4, 1.15, \infty]$. Quality of the upper right image: SNR = 27.53 dB; lower right: SNR = 27.51 dB.} \label{fig:SPF-single}
\end{figure}

A special case of the proposed method is to perform local processing with the ANN using only one FBP image. This, in fact, is a post-processing algorithm based on a regression function, which implements some non-linear local filter. In the following experiment we compare the performance of two fusion methods, one using three FBP images (sharp, normal and blurred) and another using only one FBP image produced with no low-pass filter. The results are displayed in Figure \ref{fig:SPF-single}. Visually, in the single-image fusion case some artificial streaks are observed, which do not appear in the multi-image fusion (where also a lower MSE is achieved). On the other hand, the single-image fusion produces sharper images.

%==========================================================================================================================
%==========================================================================================================================

\section{PWLS Boost - Algorithm Design and Empirical Study}\label{sect:RVTF-PWLS-it}

\subsection{Algorithm Description}

The iterative PWLS algorithm (see Section \ref{sect:BK-recon}) can be boosted by gathering intermediate versions of the image at different numbers of iterations. The idea is to capture the gradual transformation of the image from the initial to the final state. If the initial image is a blurred one, it gradually changes along the iterations towards a sharper version; the intermediate stages contain important information that can contribute to further improve the algorithm output.

The method is very similar to the one proposed in the previous section. At the training stage, a CT reconstruction is performed with a high-quality reference at hand. The examples for ANN training are produced in the following manner: the vector of inputs, corresponding to a location $q$ in the image, is assembled using neighborhoods of $q$ in the different versions of the image, gathered along the PWLS iterations. Specifically, we take a small neighborhood of pixels from each image in this sequence (see details below). The ``correct answer'', corresponding to this vector of ANN inputs, is the value of the pixel $q$ in the reference image. As was done previously, the objective function for ANN training is augmented with weights which determine the importance of the individual examples.

%==========================================================================================================================

\subsection{PWLS Boost - Empirical Study}

We conducted numerical experiments to demonstrate the proposed method using the same setup as in the FBP experiment. Training data for the ANN was obtained using a data-set of $12$ axial male thighs section images. For each, an initial image $\tilde{f}$ is computed with the FBP algorithm using a sinogram filter with cut-off frequency value of $2.0$ (see Figure \ref{fig:RV-FBP1b}). The PWLS algorithm is implemented as described in Section \ref{sect:BK-recon}, with parameters $\delta =0.02$, $\lambda=8\cdot10^{-5}$. We have performed $90$ iterations, saving an image version every $10$ iterations - overall we have a sequence of $10$ images. In practice, we use three images out of this sequence, namely those from iterations 20, 60 and 80.  From the first and the third images, neighborhoods of radius $4$ ($49$ pixels) were taken for the estimation of the pixel value, and the second image contributed a neighborhood of radius $1$ ($5$ pixels). Overall, the ANN has $2*49+5=103$ inputs. It is set to be a network with $30$ neurons in the (single) hidden layer. It has a single output, set to produce only the central pixel of the provided neighborhood.  These specific settings were obtained with a manual tuning of the design parameters.

\begin{figure}[htbp]
\centering
\begin{tabular}{l}
\includegraphics[width=1\columnwidth]{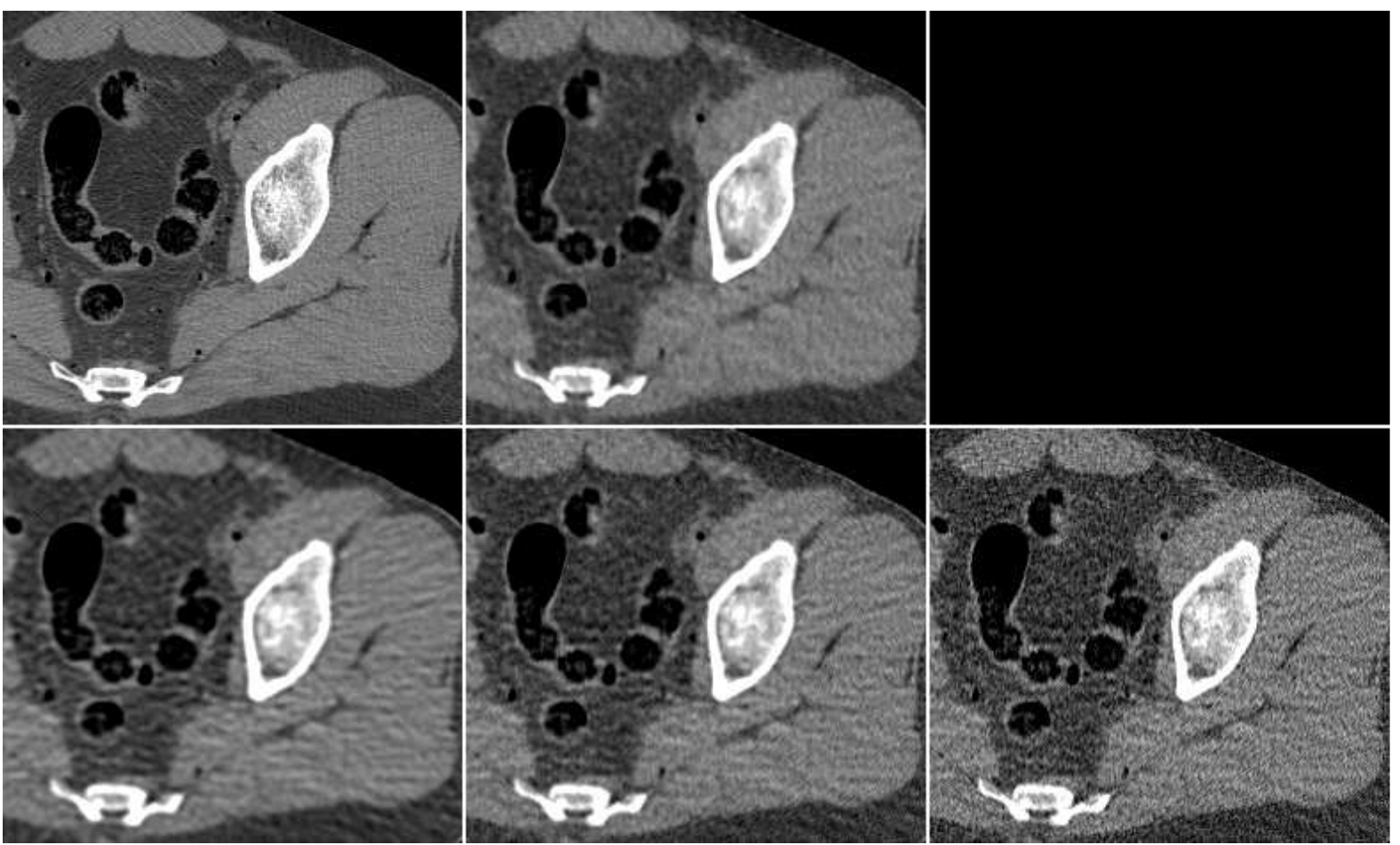}\\\centering
\includegraphics[width=1\columnwidth]{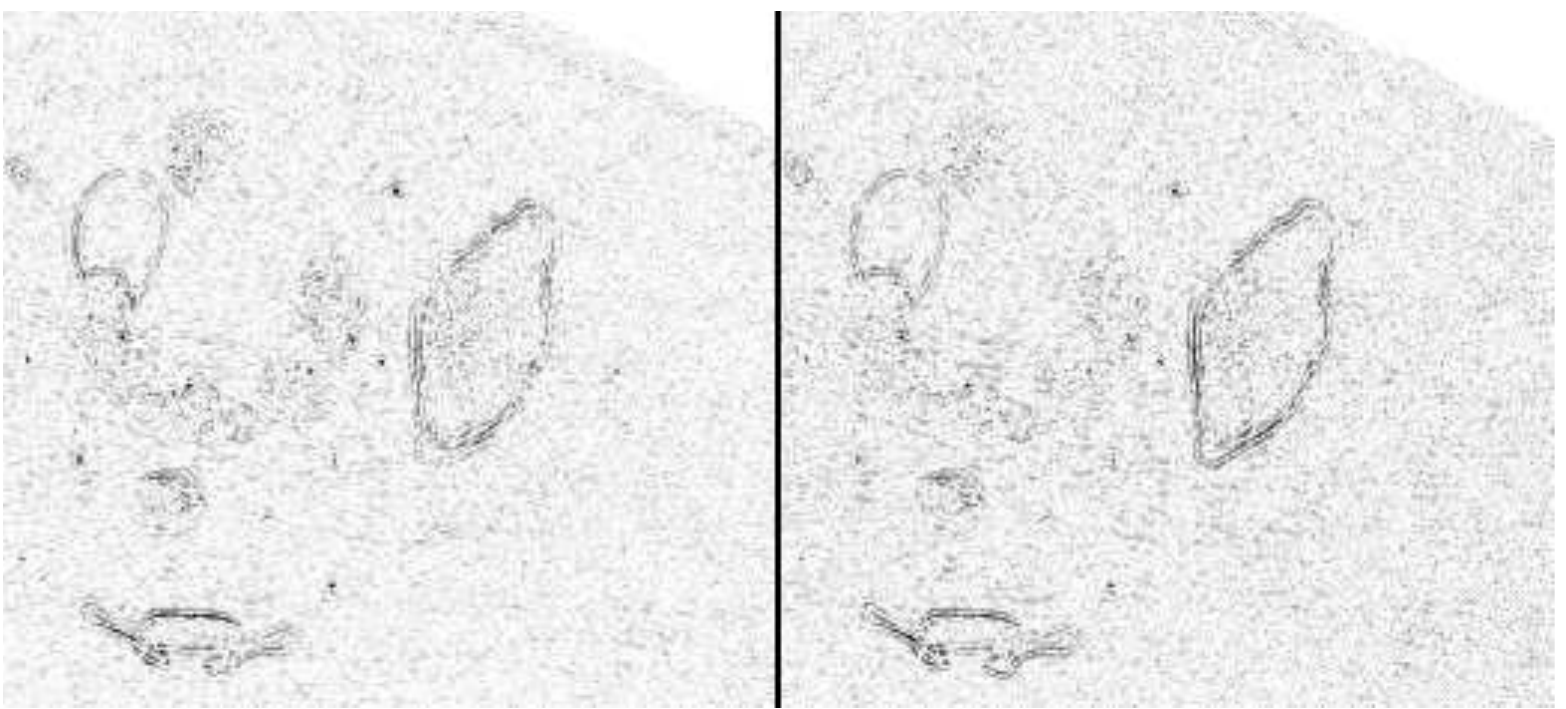}
\end{tabular}
\caption{Images of thighs section. Upper row, left to right: reference image, ANN-fused
PWLS (SNR=28.18 dB). Middle rows: three PWLS versions (20 iterations, SNR=26.05, 60 iterations,SNR=26.86 dB, 80 iterations, SNR=24.77 dB). Lower row: absolute-valued error images for the fusion image(left) and
best-SNR PWLS (right). Darker shade corresponds to a larger error.}
\label{fig:SPI-1}
\end{figure}

\begin{figure}[htbp]
\centering
\includegraphics[width=1\columnwidth]{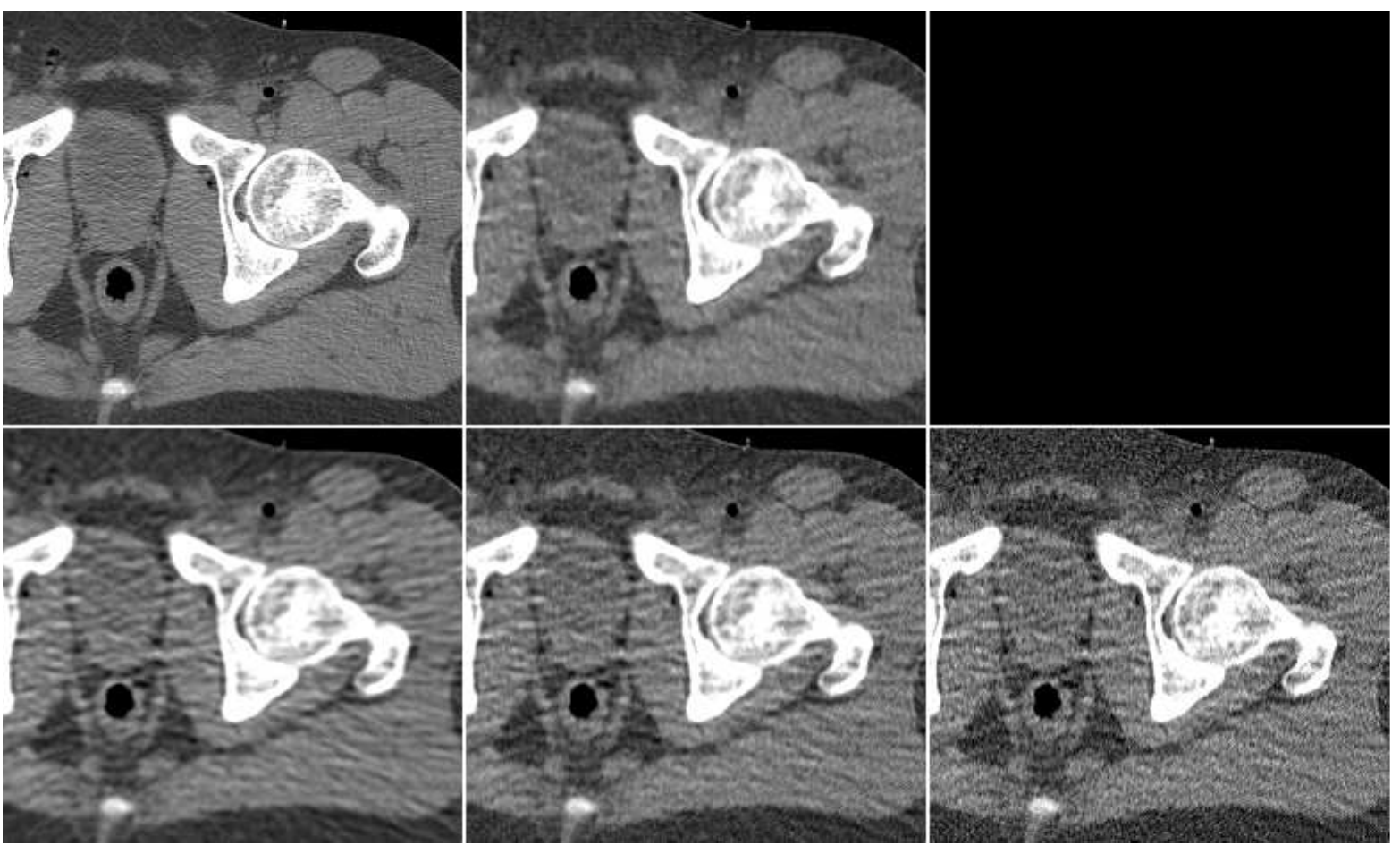}
\caption{A different test image. Left to right: reference image, ANN-fused
PWLS (SNR=27.44 dB), three PWLS versions (20 iters., SNR=25.19, 60 iters.,SNR=26.02 dB, 80 iters., SNR=24.67 dB). } \label{fig:SPI-2}
\end{figure}

In Figure \ref{fig:SPI-1} we display the fusion result along with individual PWLS reconstructions, used in the fusion process. The lower part of the figure contains the absolute-valued error images. The fusion result has a higher visual quality than any of the three underlying images. Comparing to those images, the noise level in the fusion image is the lowest, and the tissue texture is closer to the original. The sharpness is the same as in the lower middle PWLS image. The SNR values (stated in the Figure) also point to the improvement in quality. The SSIM of the fusion image is $0.95$, while the sequence of PWLS results have the SSIM values of $0.93, 0.92, 0.86$ (corresponding to the lower row of Figure \ref{fig:SPI-1}, left to right). A reconstruction of an additional test image is displayed in Figure \ref{fig:SPI-2}. The effect of the fusion observed here is similar to the one in the previous reconstruction. We conclude that the ANN-based fusion can contribute also to the iterative reconstruction, without requiring any additional iterations; the computational cost of the fusion, exercised after the reconstruction, is lower by an order of magnitude than that of the iterative process.


\begin{figure}[htbp]
\centering
\includegraphics[width=1\columnwidth]{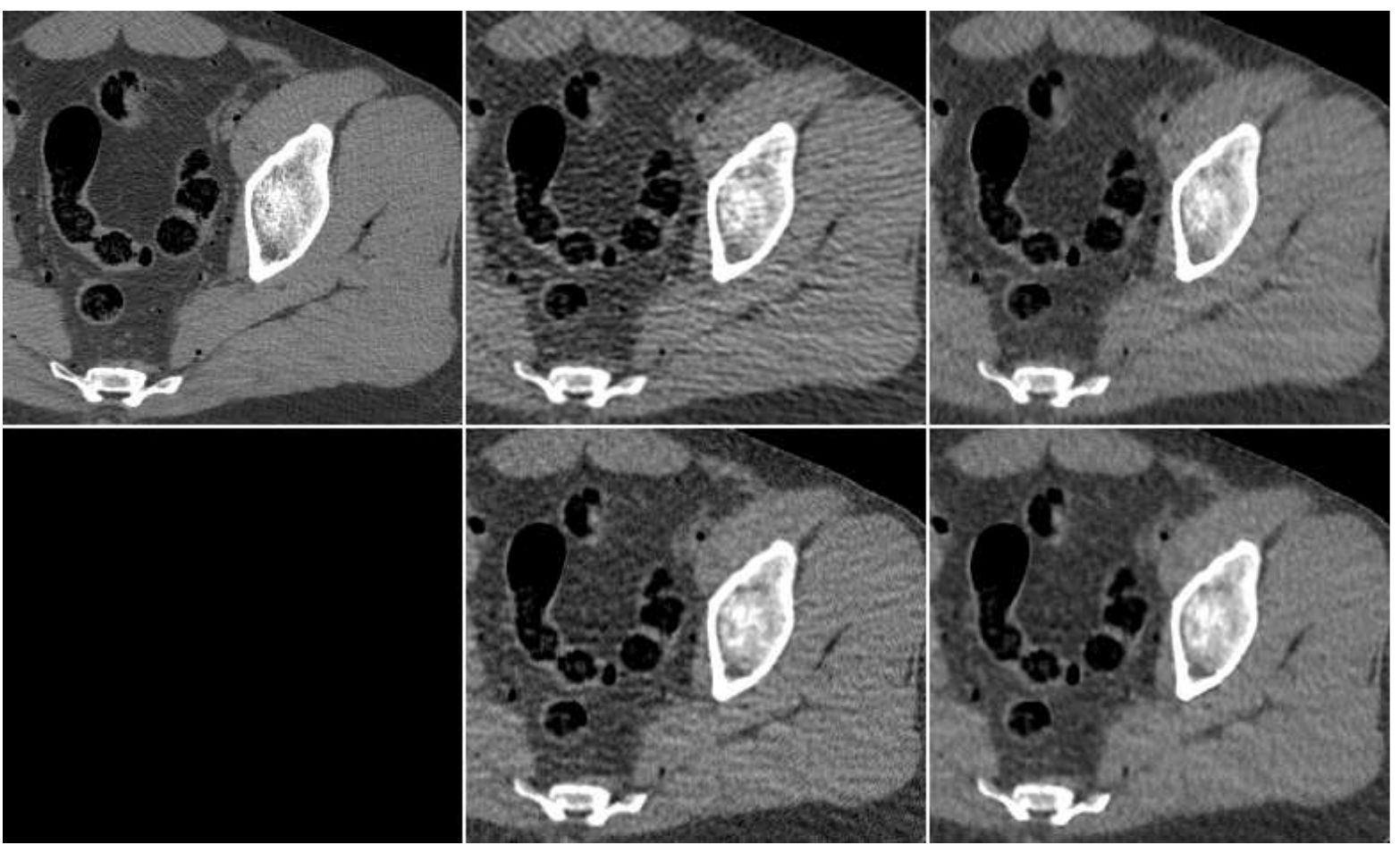}
\caption{FBP and PWLS reconstruction before (middle column) and after (right column) the ANN fusion. Upper left: reference image. Upper row: FBP (central), FBP+ANN (right). Lower row: PWLS (central), PWLS+ANN (right). } \label{fig:SPF-SPI}
\end{figure}

To summarize the fusion effect on the outcome of standard reconstruction algorithms, we display in Figure \ref{fig:SPF-SPI} images produced by both FBP and PWLS methods, before and after applying the proposed method of the ANN-based fusion; these images were previously given in Figures \ref{fig:SPF1},\ref{fig:SPI-1}.

%\noindent {\bf PROBLEM WITH A MISSING FIGURE}

% ====> the text below is irrelevant as it shows results that are not discussed in the paper. However, we may want to show results by (1) the FBP, (2) the boosted FBP, (3) the PWLS, and (4) its boosted version. In such a case, we need to form such a figure and accompany it with SNR values.

%In Figure \ref{fig:parade-all} we present a set of versions of the thighs image produced by all four of the (full-image) reconstruction algorithms presented in this paper -- . It is given . The shrinkage-based result is most sharp, but also contains streaks; this may indicate the need to adjust the weight $\mu$ of the edge-promoting component in the penalty function. The Shrinkage-based image has the lowest noise level, but the texture of the PWLS fusion result is closer to the original. Finally, the FBP fusion image is expectedly of the lowest quality among those versions, but it is reproduced almost with no the strikes typical to the FBP.
%
%\begin{figure}[htbp]
%\centering
%\includegraphics[width=1\columnwidth]{Figures/All-parade.pdf}
%\caption{Test image reconstruction by all the proposed algorithm. Upper to lower, left to right: Reference image, PWLS, Sparsity-based reconstruction, PWLS fusion, FBP fusion, Learned Shrinkage based reconstruction.} \label{fig:parade-all}
%\end{figure}

\begin{figure}[htbp]
\centering
\begin{tabular}{l}
\includegraphics[width=1\columnwidth]{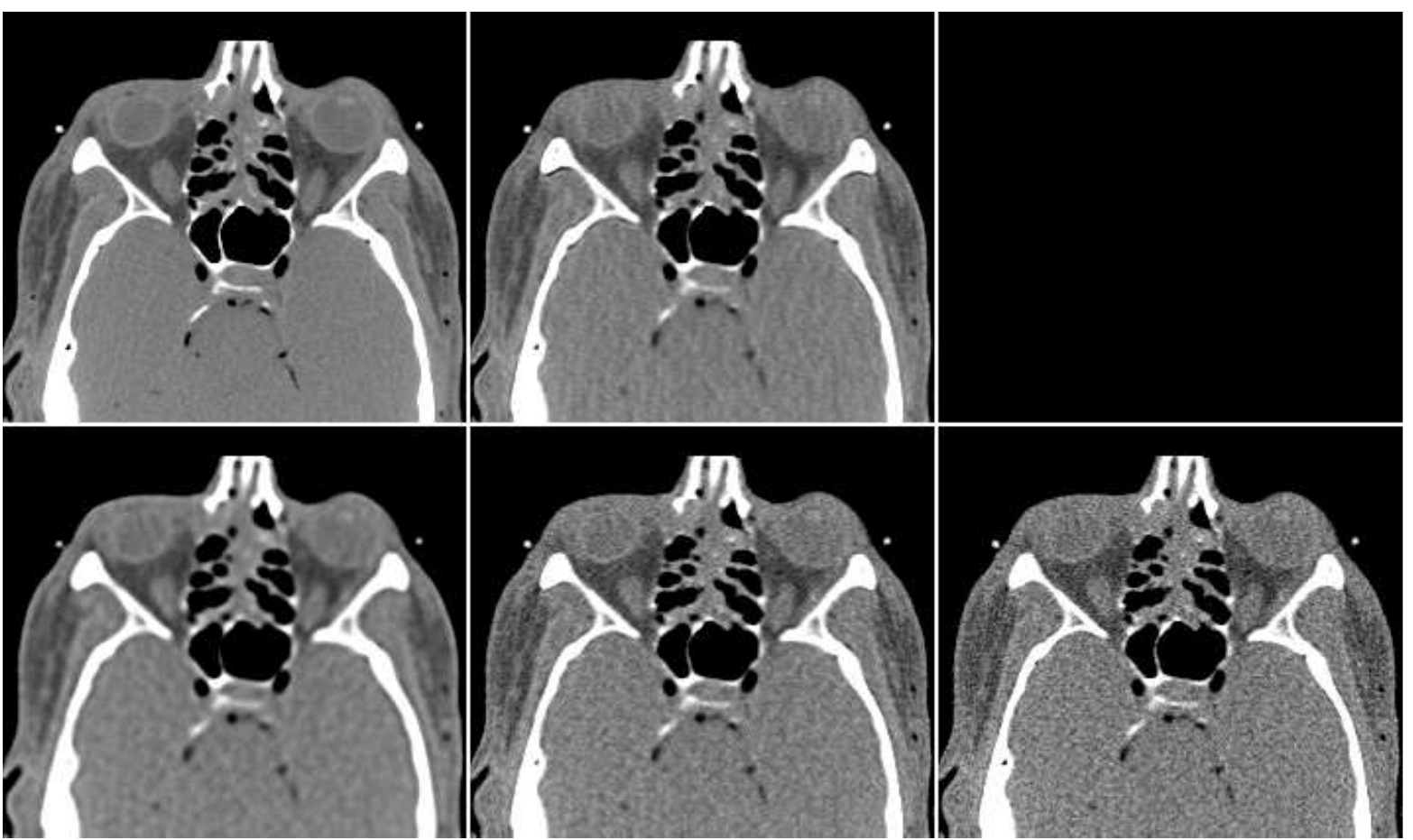}\\
\includegraphics[width=1\columnwidth]{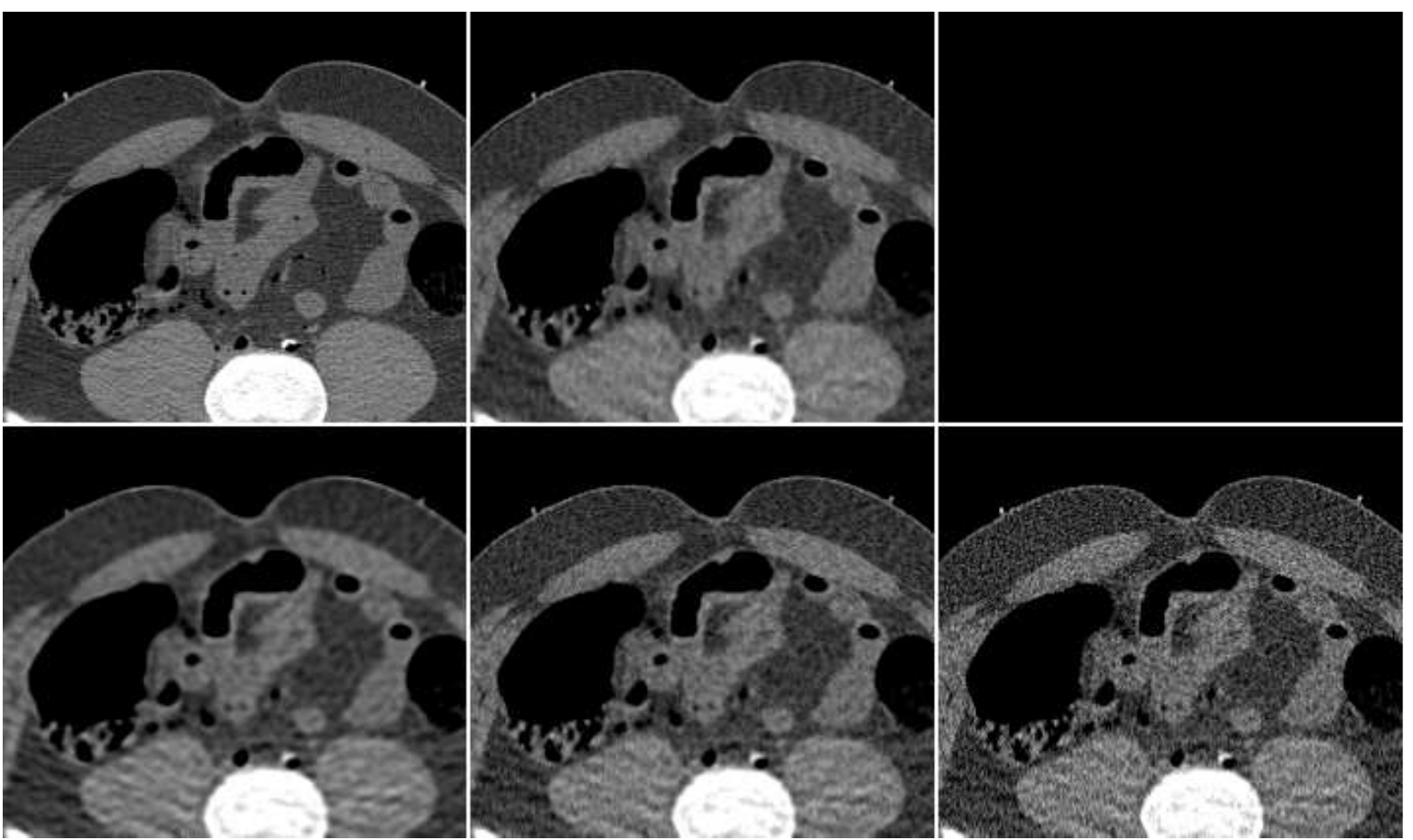}
\end{tabular}
\caption{Upper part: head reconstruction. Left to right, top to bottom: reference image, ANN-fused
image, three PWLS versions (20, 60, 80 iterations). Lower part: abdomen reconstruction with the same arrangement.}
\label{fig:SPI-3}
\end{figure}

In order to test the robustness of the training results, we apply the ANN trained with the thigh sections, for a reconstruction of images of other body parts -- sections of the head and the abdomen. Reconstruction results are presented in Figure \ref{fig:SPI-3} in the same order as in the previous comparison: middle image in the upper row is the result of fusion, which components are presented in the lower row. The head reconstruction is improved substantially by the fusion process, as visual observation shows. However, the SNR values (given in Table \ref{table:SPI-anatomic}) point to the favor of the PWLS image corresponding to $60$ iterations (lower middle image). The highest SSIM value does belong to the fusion result, though. In the case of the abdomen section, the fusion image is similar to the $40$-iterations version but contains less noise; its quantitative measures are somewhat better than those of the individual PWLS images.

\begin{table}[htbp]
\begin{center}
%\scriptsize{
\begin{tabular}{||l|l|l|l|l||}
\hline \hline

Image   & PWLS &PWLS & PWLS  & Fusion \\
        & $40$ iters. &$60$ iters. & $20$ iters.& result \\
\hline \hline
Head section &&&&\\
\hline
SNR (plain)  &  24.91 &  28.67 & 28.12 &  28.09\\
\hline
SSIM  &  0.878 &  0.873  &  0.858  &  0.881\\
 \hline  \hline

Abdomen section &&&&\\
\hline
SNR (plain)  &  26.68 &  27.15 & 25.24 &  27.94\\
\hline
SSIM  &  0.813 &  0.800  &  0.761  &  0.821\\
 \hline  \hline
\end{tabular}
\end{center}\caption{Quantitative measures for the head and abdomen reconstructions.}
\label{table:SPI-anatomic}
\end{table}

\begin{figure}[htbp]
\centering
\includegraphics[width=1\columnwidth]{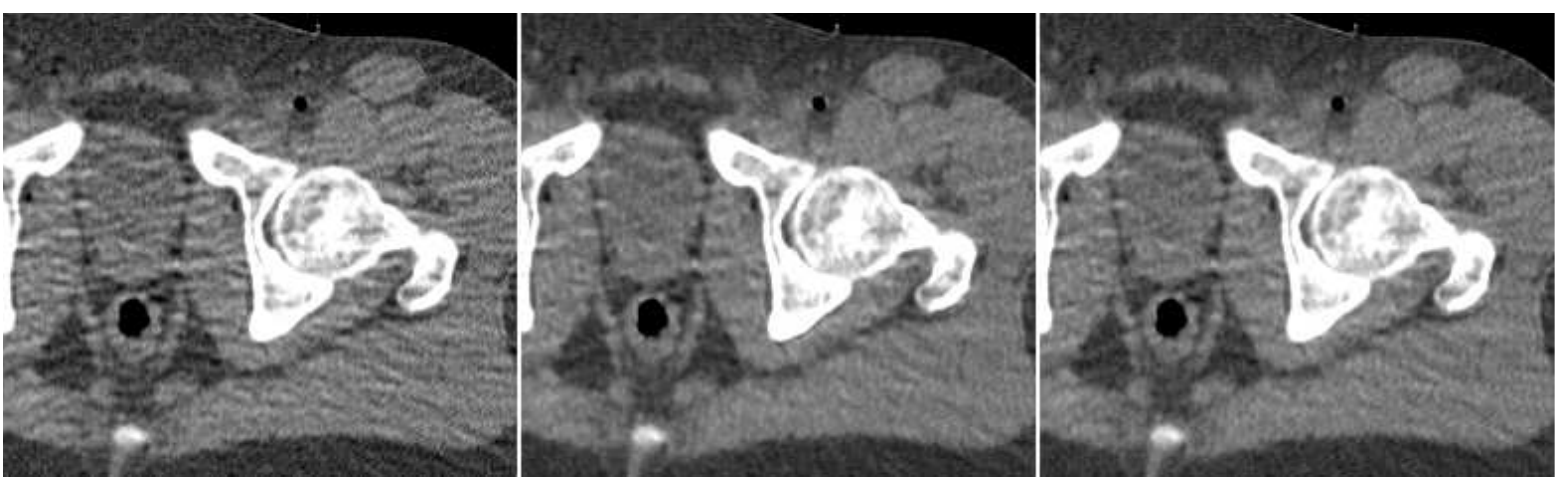}
\caption{Left to right: PWLS image (60 iterations, SNR=26.02 dB), single-image fusion (SNR = 27.25 dB), multi-image fusion (SNR = 27.44 dB). } \label{fig:SPI-single}
\end{figure}

As a last experiment, we consider the special case where the ANN only performs a local filtering of the single version of the image, without a reference to the other versions. A neighborhood of radius $r=4$ ($49$ pixels) was extracted for each location in the PWLS image, corresponding to iteration number $60$. The fusion result is visually compared in Figure \ref{fig:SPI-single} versus the image produced from $10$ PWLS versions, as before. It can be observed that the processing by ANN reduces the noise appearing in the PWLS image, but it is slightly inferior to the fusion image produced from several PWLS versions.

%==========================================================================================================================
%==========================================================================================================================

\section{Computational Complexity of The Method}\label{sect:complex}

We analyze the number of computations required for the proposed method in the cases of FBP and PWLS reconstruction. First, we consider the complexity of applying the ANN to perform the pixel-wise fusion from a number of image versions produced at the reconstruction stage. For an $n\times n$ image, $n^2$ activations of the ANN is required. Typically, the dimension of the input vector of the ANN is of the order of $100$ samples, the output dimension has up to $5$ elements, and a single hidden layer of up to $40$ neurons is used. Thus, the network contains $40\cdot100+5\cdot 40 = 4200$ weights\footnote{This number can be reduced if a parallel implementation of the ANN is available, since each neuron output can be calculated  separately}. Each neuron also implements a sigmoid function thus requiring $40$ sigmoid calculations to produce the ANN output values. Therefore, the cost of performing a local fusion by the ANN is ~$4200\cdot n^3$ operations.

When the method is used for the FBP reconstruction, a number of FBP versions must be produced; in our experiments, three reconstructions suffice. Each FBP reconstruction is of computational complexity of $\mc{O}(n^3)$. Therefore, if no hardware changes in an existing scanner are made, producing the fusion image will require roughly four times the extent of a single reconstruction (three FBP processes and the fusion step). Of course, the regular FBP image will be available for the radiologist after the usual time of a single FBP reconstruction.

As for the iterative PWLS algorithm, no changes in the reconstruction process are needed, since we only sample images along the standard iterations. We do not have an accurate estimate for the time complexity of the PWLS, since it depends on the optimization method and its efficient implementation. However, the iterative process necessarily involves an application of the system matrix ($\mc{O}(n^3)$ operations) in each iteration, and therefore it is by order of magnitude slower than the FBP. Adding the fusion step in the end of this process will only marginally increase the total reconstruction time.

%==========================================================================================================================
%==========================================================================================================================

\section{Summary}\label{sect:summary}

We have introduced a method for quality improvement for a general parametric signal estimator. The concept is to use a regression function for a local fusion of a number of estimator's outputs, corresponding to different parameter settings. The regression proposed is realized with feed-forward artificial neural networks. The fusion process consists of two components: first, the behavior of the signal in its different versions is gathered; second, the ANN performs its own non-linear filtering of the signal versions in small neighborhoods of the estimated pixel.

The proposed method is very general and CT reconstruction is only one possible application for it. The local fusion can be used to solve any linear on non-linear inverse problem where an algorithm, producing a solution estimate, exists. The proposed method will enable to incorporate the algorithm outputs, produced with different values of a core parameter, to a single improved result, thus removing the need for tuning this parameter.

In this work this concept was illustrated for the case of CT reconstruction, done with two basic algorithms -- the FBP and the PWLS. Empirical results suggest that the local fusion can improve on the resolution variance trade-off of the given reconstruction algorithm, thus adding to the visual quality of the CT images. The post-processing method is not very time-consuming, and  the cost of the local fusion can be  well below the extent of one FBP reconstruction.

%==========================================================================================================================
%==========================================================================================================================

\bibliographystyle{IEEEbib}
\bibliography{ct_neural_net_arxive}
%\bibliography{IR_bib}
\end{document}